%% file: ms.tex
\documentclass[10pt]{article}
\usepackage{times}
\usepackage{helvet}
\usepackage{courier}
\usepackage{url} 
\usepackage{graphicx}
\usepackage{graphicx}
\usepackage{caption} %
\usepackage[switch]{lineno}
\usepackage{mathtools}
\usepackage{amsmath,environ}
\usepackage{bbm}
\usepackage{amssymb}
\usepackage{amsthm}
\usepackage{algorithmicx}
\usepackage{algorithm}
\usepackage{algpseudocode}
\newtheorem{definition}{Definition}
\newtheorem{lemma}{Lemma}
\newtheorem{theorem}{Theorem}
\newtheorem{corollary}{Corollary}
\newtheorem{remark}{Remark}
\newcommand{\norm}[1]{\left\lVert #1 \right\rVert}
\newcommand{\norms}[1]{\lVert #1 \rVert}
\newcommand{\inner}[2]{\langle #1, #2 \rangle}
\newcommand{\lep}[1]{\mathop  \le \limits^{(#1)}}

\newcommand{\ep}[1]{\mathop  = \limits^{(#1)}}
\newcommand{\ex}[1]{\mathbb{E}\left[ #1 \right] }
\newcommand{\exm}[1]{\mathbb{E}_m\left[ #1\right]}
\newcommand{\exo}[1]{\mathbb{E}_0\left[ #1\right]}
\newcommand{\mh}{\mathcal{H}} 
\DeclareMathOperator*{\argmax}{arg\,max}
\graphicspath{{figures_CI/}}
\usepackage{graphicx, caption, subcaption, wrapfig}

\NewEnviron{lmath}{%
\begin{linenomath}\begin{align}
  \BODY
\end{align}\end{linenomath}
}
\NewEnviron{lmath*}{%
\begin{linenomath}\begin{align*}
  \BODY
\end{align*}\end{linenomath}
}
\NewEnviron{lm}{%
\begin{linenomath} \BODY \end{linenomath}
}

\title{Local Differential Privacy for Bayesian Optimization}
\author{Xingyu Zhou \thanks{This work was done during the internship of the first author at Alibaba, USA.}\\Department of ECE\\The Ohio State University\\\url{zhou.2055@osu.edu} \and Jian Tan\\
Alibaba, USA\\\url{j.tan@alibaba-inc.com}\\} 

\begin{document}
\urlstyle{tt}
 	\maketitle
  	\begin{abstract}
  		Motivated by the increasing concern about privacy in nowadays data-intensive online learning systems, we consider a black-box optimization in the nonparametric Gaussian process setting with local differential privacy (LDP) guarantee. Specifically, the rewards from each user are further corrupted to protect privacy and the learner only has access to the corrupted rewards to minimize the regret. We first derive the regret lower bounds for any LDP mechanism and any learning algorithm. Then, we present three almost optimal algorithms based on the GP-UCB framework and Laplace DP mechanism. In this process, we also propose a new Bayesian optimization (BO) method (called MoMA-GP-UCB) based on \underline{m}edian-\underline{o}f-\underline{m}eans techniques and kernel \underline{a}pproximations, which complements previous BO algorithms for heavy-tailed payoffs with a reduced complexity. Further, empirical comparisons of different algorithms on both synthetic and real-world datasets highlight the superior performance of MoMA-GP-UCB in both private and non-private scenarios.
  	\end{abstract}

  \input{intro}
	\input{model}
	\input{main}
\input{simulations}
  \section{Conclusion}
    We derived regret lower bounds for LDP BO and presented three almost optimal algorithms. We also proposed MoMA-GP-UCB. It complements previous BO algorithms for heavy-tailed payoffs and has superior performance with a reduced complexity.

	\bibliographystyle{plain}
	\bibliography{ref}

  \appendix
  \input{appendix}

\end{document}

%% file: intro.tex
\section{Introduction}
We consider the problem of maximizing an unknown function $f$ over a set $\mathcal{D}$ via sequentially querying it and received only bandit feedback, i.e., when we query at $x$, we observe a possibly noisy evaluation of $f(x)$. This model has been a main focus in machine learning research, e.g., the classic multi-armed bandit (MAB) setting~\cite{lai1985asymptotically}, linear bandit setting~\cite{abbasi2011improved} and the general Bayesian optimization~\cite{shahriari2015taking}, with each one generalizing the previous one. It also finds broad applications in many real-world systems, including medical experiments, online shopping websites and recommender systems~\cite{li2010contextual}. These systems adaptively make a decision and receive rewards (feedback) from the user to simultaneously learn insightful facts and maximize the profit.

Recently, privacy has become a key issue in the above mentioned online learning systems. Users have become increasingly concerned about directly sharing their online information or activities to these systems, since these activities may reveal their private information. For example, a customer of an online shopping website is not willing to tell the website that he or she has purchased medicines for mental issues. Another example is the medical experiments in which the patient could reject to share the actual effects of the treatment due to privacy concerns. This stimulates the need to have a mechanism that further corrupts the feedbacks from each user to protect privacy, which exactly fits the \emph{locally differential private} (LDP) model~\cite{kasiviswanathan2011can,duchi2013local}.

In contrast to the standard differential privacy model~\cite{dwork2014algorithmic}, in which the learner collects the true data while releasing a private output to protect privacy, in the LDP model, the learner only has access to corrupted input data from the users. Hence, LDP often provides a much stronger privacy protection for the user and is more appealing in real applications, especially for the systems mentioned above~\cite{cormode2018privacy}. To the best of our knowledge, in the setting of online learning with bandit feedback, LDP model has only been studied theoretically very recently. For example, in~\cite{gajane2018corrupt,ren2020multi}, the authors investigate MAB with LDP guarantee. \cite{zheng2020locally} studies LDP in the linear (contextual) bandit setting. However, LDP in the most general scenario, i.e., Bayesian optimization (BO), remains an important open problem.

Motivated by this, in this paper, we investigate the locally differentially private BO, in which the rewards are further corrupted to protect privacy. Specifically, we consider a Gaussian process (GP) model for BO (also called Gaussian process bandit setting), which directly generalizes both MAB and linear bandit setting. The main contributions of this paper can be summarized as follows.

\textbf{Contributions.} (i) We first derive the regret lower bounds for any LDP mechanism and any learning algorithm. (ii) Then, we present three almost optimal algorithms based on the GP-UCB framework and Laplace DP mechanism. (iii) Our two new methods developed for handling LDP also contribute to BO under heavy-tailed payoffs in general. In particular, one is a new truncation method that can be applied to any sub-Weibull rewards. The other one,  called MoMA-GP-UCB, is based on median-of-means techniques and kernel approximations, which complements previous BO algorithms for general heavy-tailed payoffs~\cite{chowdhury2019bayesian} with a reduced complexity. (iv) We further conduct empirical comparisons of different algorithms over both synthetic and real-world datasets, which demonstrate the superior performance of our new MoMA-GP-UCB algorithm in both private and non-private settings.

\subsection{Related Work}
In the traditional non-private case, a line of BO methods based on Gaussian process (GP) and upper confidence bound (UCB) have been analyzed in both sub-Gaussian~\cite{srinivas2009gaussian,chowdhury2017kernelized} and heavy-tailed scenarios~\cite{chowdhury2019bayesian}. Kernel approximation is recently proposed to reduce complexity of GP-based BO algorithms~\cite{mutny2018efficient,calandriello2019gaussian}. In the private BO case,~\cite{kusner2015differentially} studies how to privately release the BO outputs to protect privacy (e.g., hyper-parameters of machine learning model), and hence it belongs to the traditional DP perspective rather than LDP.

LDP model has been previously considered in MAB setting~\cite{gajane2018corrupt,basu2019differential,ren2020multi}. Recently, it is generalized to linear contextual bandits in which both the rewards and the contexts are corrupted for privacy~\cite{zheng2020locally}. There are also some other types of DP considered in MAB and linear bandit setting (not comparable to our work). Due to space limitations, we refer readers to~\cite{ren2020multi,zheng2020locally} and the references therein.

%% file: model.tex
\section{Problem Statement and Preliminaries}
	We consider a sequential decision-making problem over a set $\mathcal{D}$. A learning policy is adopted to select an action $x_t \in \mathcal{D}$ at each discrete time slot $t=1,2,\ldots$ with the corresponding reward observation $y_t = f(x_t) + \eta_t$, i.e., $y_t$ could be a noisy version of $f(x_t)$. Then, the reward $y_t$ will be further corrupted to protect privacy, and only the private response $\bar{y}_t$ is revealed to the learning agent. The action $x_t$ is chosen based on the arms played and the private rewards obtained until $t-1$, denoted by the history $\mathcal{H}_{t-1} =  \{(x_s,\bar{y}_s): s\in [t-1]\footnote{For any positive integer, we define [m]:=\{1,2, \ldots, m\}} \}$.  The objective is to simultaneously preserve LDP and minimize the cumulative regret defined as 
	\begin{linenomath}
		\begin{align}
		R_T = \sum_{t=1}^T f(x^*) - f(x_t),
	\end{align}
	\end{linenomath}
	where $x^* = \arg\max_{x\in \mathcal{D}}f(x)$ (assuming the maximum is attained).
	\begin{definition}[$(\epsilon,\delta)$-LDP]
		A randomized mechanism $\mathcal{Q}: \mathcal{D} \to \mathcal{Z}$ is said to protect $(\epsilon,\delta)$-LDP if for any $x,x'\in\mathcal{D}$, and any measurable subset $E \in \mathcal{Z}$, there is
		\begin{linenomath}
			\begin{align*}
			\mathbb{P}\{ M(x) \in E\} \le e^{\epsilon}\mathbb{P}\{M(x')\in E\} + \delta,
		\end{align*}
		\end{linenomath}
		for $\epsilon \ge 0$ and $\delta \ge 0$. Moreover, if $\delta =0$, we say it protects $\epsilon$-LDP.
	\end{definition}
	Note that, if not explicitly stated, LDP in this paper means $\epsilon$-LDP (stronger than $(\epsilon,\delta)$-LDP).

	\textbf{Noise Assumptions.} We assume that the noise $\eta_t$ has zero mean conditioned on the history and is bounded by $R$ almost surely. We also address the case of unbounded noise at the end of the paper.

	\textbf{Regularity Assumptions.} Attaining a sub-linear regret is in general infeasible for an arbitrary reward function $f$ over a very large space without any assumptions on the structure of $f$. In this paper, we assume that $\mathcal{D}$ is compact and $f$ has a bounded norm in the RKHS of functions $\mathcal{D} \to \mathbb{R}$, corresponding a kernel function $k: \mathcal{D} \times \mathcal{D} \to \mathbb{R}$. This RKHS denoted by $\mathcal{H}_k(\mathcal{D})$ is completely determined by its kernel function with an inner product $\inner{\cdot}{\cdot}_{\mathcal{H}}$ that satisfies the reproducing property: $f(x) = \inner{f}{k(x,\cdot)}_{\mh}$ for all $f \in H_k(\mathcal{D})$. The norm for the RKHS is given by $\norm{f}_{\mathcal{H}} = \sqrt{\inner{f}{f}_{\mathcal{H}}}$, which measures the smoothness of $f$. We assume $\norm{f}_{\mathcal{H}} \le B$ and $B < \infty$ is a known constant. Moreover, we assume a bounded variance by restricting $k(x,x) \le 1$. Note that two commonly used kernels \emph{Squared Exponential} and \emph{Mat\'ern} satisfy the bounded variance assumption, defined as:
	\begin{linenomath}
		\begin{align*}
		k_{\text{SE} }(x,x') &= \exp{(-s^2/2l^2)}\\
		k_{\text{Mat\'ern} }(x,x') &= \frac{2^{1-\nu}}{\Gamma(\nu)}\left(\frac{s\sqrt{2\nu}}{l}\right)^{\nu}B_{\nu}\left(\frac{s\sqrt{2\nu}}{l}\right),
	\end{align*}
	\end{linenomath}
	where $l > 0$ and $\nu > 0$ are hyper-parameters, $s = \norm{x - x'}_2$ specifies the similarity between two points, and $B_{\nu}(\cdot)$ is the modified Bessel function.

	\textbf{Surrogate GP Model\footnote{The surrogate GP model described above (i.e., a GP prior and a Gaussian likelihood) is only used for the algorithm design.}.} A Gaussian process, denoted by $\mathcal{GP}(\mu(\cdot),k(\cdot,\cdot))$, is a collection of (possibly infinitely many) random variables $f(x), x\in \mathcal{D}$, such that every finite subset of random variables $\{f(x_i), i\in [m] \}$ is jointly Gaussian with mean $\ex{f(x_i)} = \mu(x_i)$ and covariance $\ex{(f(x_i)-\mu(x_i))(f(x_j) - \mu(x_j))} = k(x_i, x_j)$, where $i,j \in [m]$ and $m\in \mathbb{N}$. By conditioning GPs on available observations, one can obtain a non-parametric surrogate Bayesian model over the space of functions. In particular, we use $\mathcal{GP}(0,k(\cdot,\cdot))$ as an initial prior on the unknown black-box function $f$, and a Gaussian likelihood with the noise variables $\eta_t$ drawn independently across $t$ from $\mathcal{N}(0,\lambda)$. Conditioned on a set of past observations $\mathcal{H}_t = \{(x_s,{y}_s), s\in[t] \}$, by the properties of GPs~\cite{rasmussen2003gaussian}, the posterior distribution over $f$ is $\mathcal{GP}(\mu_t(\cdot),k_t(\cdot,\cdot))$, where 
	\begin{linenomath}
		\begin{align}
		\mu_t(x) &= k_t(x)^T(K_t + \lambda I)^{-1}{y}_{1:t}\label{eq:mu}\\
		k_t(x,x')&=k(x,x')-k_t(x)^T(K_t + \lambda I)^{-1}k_t(x')\nonumber\\
		\sigma_t^2(x) &= k_t(x,x)\label{eq:sigma}.
	\end{align}
	\end{linenomath}
	Therefore, for every $x\in\mathcal{D}$, the posterior distribution of $f(x)$, given $\mathcal{H}_t$ is $\mathcal{N}(\mu_t(x),\sigma_t^2(x))$. The following term often plays a key role in the regret bounds of GP based algorithms.
	\begin{linenomath}
		\begin{align*}
		\gamma_t :=\gamma_t(k,\mathcal{D}) = \max_{A \subset \mathcal{D}: |A| = t} \frac{1}{2}\ln |I_t + {\lambda}^{-1}K_A |,
	\end{align*}
	\end{linenomath}
	where $K_A = [k(x,x')]_{x,x'\in A}$. Roughly speaking, $\gamma_t$ is the maximum mutual information that can be obtained about the GP prior from $t$ samples corrupted by a Gaussian channel $\mathcal{N}(0,\lambda)$. It is a function of the kernel $k$ and domain $\mathcal{D}$. For instance, if $\mathcal{D}$ is compact and convex, then we have $\gamma_t = O((\ln t)^{d+1})$ for $k_\text{SE}$, $O(t^{\frac{d(d+1)}{2\nu+d(d+1)}}\ln t)$ for $k_{\text{Mat\'ern}}$, and $O(d\ln t)$ for a linear kernel~\cite{srinivas2009gaussian}.


%% file: main.tex
\section{Lower Bounds}
In this section, we derive the lower bounds for both $k_\text{SE}$ and $k_{\text{Mat\'ern}}$ under any LDP mechanism and any learning algorithm, as presented in the following theorem.
\begin{theorem}
\label{thm:lb}
	Let $\mathcal{D} = [0,1]^d$ for some $d \in \mathbb{N}$. Fix a kernel $k \in \{k_{\text{SE} },k_{\text{Mat\'ern} }\}$, $B>0$, $\epsilon>0$, $T \in \mathbb{Z}$, $\delta \in (0,1)$, $\alpha \in (0,1]$ and $v > 0$. Given any learning algorithm, any $\epsilon$-LDP mechanism, there exists a function $f\in \mathcal{H}_k(\mathcal{D})$ with $\norm{f}_{\mathcal{H}} \le B$, and a reward distribution satisfying $\ex{|y_t|^{1+\alpha} | \mathcal{F}_{t-1}} \le v$ for all $t \in [T]$, such that the following hold, respectively
	\begin{itemize}
            \item $\ex{R_T} = \Omega\left(v^{\frac{1}{1+\alpha}}T^{\frac{1}{1+\alpha}} {\zeta^{\frac{-2\alpha}{1+\alpha}}} \left(\ln\frac{B^{(1+\alpha)/\alpha}T\zeta^2 }{v^{1/\alpha}} \right)^{\frac{d\alpha}{2+2\alpha}} \right)$, where $\zeta = e^{\epsilon}-1$, if $k = k_{\text{SE}}$ 

            \item $\ex{R_T} = \Omega\left( v^{\frac{\nu}{\nu(1+\alpha)+d\alpha}}T^{\frac{\nu+d\alpha}{\nu(1+\alpha)+d\alpha}} \tilde{\zeta}  B^{\frac{d\alpha}{\nu(1+\alpha)+d\alpha}}  \right)$, where $\tilde{\zeta} = {\zeta^{\frac{-2\alpha}{1+\alpha}+{\frac{2d\alpha^2}{(1+\alpha)(\nu(1+\alpha)+d\alpha)} }} }$ and $\zeta = e^{\epsilon}-1$, if $k = k_{\text{Mat\'ern} }$.
        \end{itemize}
\end{theorem}

\begin{remark} For a small $\epsilon$ and $\alpha = 1$, and hence $\zeta \approx \epsilon$, the regret lower bounds in Theorem~\ref{thm:lb} have an additional factor of $1/\epsilon$ in front of the lower bounds for non-private case in~\cite{chowdhury2019bayesian}\footnote{for $k = k_{\text{Mat\'ern} }$, it holds for a large $\nu$.}.
\end{remark}

\begin{proof}[Proof Sketch of Theorem~\ref{thm:lb}]
	The proof follows the standard techniques in~\cite{scarlett2017lower,chowdhury2019bayesian}, which provide lower bounds for \emph{non-private} BO under $i.i.d$ Gaussian noise (or heavy-tailed payoffs). The key challenge is handle the additional requirement of $\epsilon$-LDP. To this end, we aim to relate the Kullback-Leibler (KL) divergence between two distributions $P_1$ and $P_2$ to the KL divergence between two new distributions $M_1$ and $M_2$, which are the distributions transformed from $P_1$ and $P_2$ according to a given $\epsilon$-LDP mechanism. Inspired by~\cite{basu2019differential}, we resort to Theorem 1 of~\cite{duchi2013local} and Pinsker’s inequality. More specifically, by Theorem 1 of~\cite{duchi2013local}, we have 
	\begin{linenomath}
		\begin{align*}
		D_{kl}(M_1 || M_2) + D_{kl}(M_2 || M_1) &\le 4(e^{\epsilon}-1)^2||P_1 - P_2 ||^2_{TV}.
	\end{align*}
	\end{linenomath}
	Then, by Pinsker’s inequality, we have 
	\begin{linenomath}
		\begin{align*}
		||P_1 - P_2 ||^2_{TV} \le 2D_{kl}(P_1 || P_2).
	\end{align*}
	\end{linenomath}
	Thus, roughly speaking,  there is an additional term $(e^{\epsilon}-1)^2$. The full proof is in Appendix.
\end{proof}

\section{Algorithms and Upper Bounds}
In this section, we will present three algorithms that are able to achieve nearly optimal regret while guaranteeing $\epsilon$-LDP. All the three algorithms rely on adding additional Laplace noise on the reward (i.e., Laplace mechanism in DP) to provide privacy guarantee. Note that, due to the additional Laplace noise, the rewards received by the learner are now no longer sub-Gaussian, and hence standard algorithms will not work. As a result, the three algorithms mainly differ in the way of handling the issue of non-sub-Gaussian rewards.

\subsection{Laplace Mechanism}

A commonly used mechanism in the areas of DP is the Laplace mechanism, which adds independent Laplace noise to the data point. For any $\mathcal{L}>0$, the PDF of the Laplace($\mathcal{L}$) (i.e., mean is zero) is given by
\begin{linenomath}
	\begin{align*}
	\text{Laplace}(\mathcal{L}): l(x\mid \mathcal{L}) = (2\mathcal{L})^{-1}\exp(-|x|/\mathcal{L}).
\end{align*}
\end{linenomath}
Thus, it is with mean $0$ and variance $2\mathcal{L}^2$. The Laplace mechanism used in this paper is stated in Curator~\ref{CTL} and its theoretical guarantee is given by Lemma~\ref{lem:CTL}.

\makeatletter
\renewcommand*{\ALG@name}{Curator}
\makeatother
\begin{algorithm}
\caption{Convert-to-Laplace (CTL($\epsilon$))}\label{CTL}
\textbf{On receiving} a reward observation $y_t$:
		\begin{algorithmic}
			\State \Return $\bar{y}_t :=y_t + L$, where $L\sim$ Laplace$(\mathcal{L})$ and $\mathcal{L} = \frac{2(B+R)}{\epsilon}$.
		\end{algorithmic}
\end{algorithm}

\begin{lemma}
\label{lem:CTL}
	CTL($\epsilon$) guarantees $\epsilon$-LDP.
\end{lemma}
\begin{proof}
	See Appendix.
\end{proof}

\subsection{Adaptively Truncated Approximate (ATA) Algorithm and Regret}
One direct way of handling non-sub-Gaussian rewards in BO is to utilize the recently developed technique for heavy-tailed payoffs~\cite{chowdhury2019bayesian}. In particular, the authors show that when combining a good feature approximation (e.g., Nystr\"{o}m approximation) and a feature adaptive truncation of rewards (e.g., TOFU in~\cite{shao2018almost}), one can obtain a regret bound roughly $\tilde{O}(\gamma_T T^{\frac{1}{1+\alpha}})$, when the $(1+\alpha)$-th moment of the reward is finite and $\alpha \in (0,1]$. Hence, when $\alpha = 1$, it recovers the regret bounds under sub-Gaussian rewards~\cite{chowdhury2017kernelized}.

Thus, it is natural to adapt ATA-GP-UCB introduced in~\cite{chowdhury2019bayesian} to handle the non-sub-Gaussian payoffs caused by the Laplace noise in the LDP setting, which leads to the LDP-ATA-GP-UCB, as described in Algorithm~\ref{alg:ATA-GP-UCB}.

\setcounter{algorithm}{0}
\makeatletter
\renewcommand*{\ALG@name}{Algorithm}
\makeatother
\begin{algorithm}
\caption{LDP-ATA-GP-UCB}\label{alg:ATA-GP-UCB}
\begin{algorithmic}[1]
\State \textbf{Input:} Parameters $\lambda$, $B$, $R$, $\epsilon>0$, $\{b_t\}_{t\ge1}$, $\{\beta_t\}_{t\ge1}$, and $q$.
\State \textbf{Set:} $\tilde{\mu}_0(x) = 0$ and $\tilde{\sigma}_0(x) = k(x,x)$ for all $x \in \mathcal{D}$.
\For{$t = 1,2,3,\ldots, T$}
\State Play $x_t = \argmax_{x\in\mathcal{D}}\tilde{\mu}_{t-1}(x) + \beta_{t}(x)\tilde{\sigma}_{t-1}(x)$
\State Receive private response $\bar{y}_t$ from CTL($\epsilon$)
\State Set $\tilde{\varphi}_t(x) = \text{Nystr\"{o}mEmbedding}(\{(x_i,\tilde{\sigma}_{t-1}(x_i))\}_{i=1}^t,q)$ 
\State Set $m_t$ as the dimension of $\tilde{\varphi}_t$
\State Set $\tilde{\Phi}_t^T = [\tilde{\varphi}_t(x_1), \ldots, \tilde{\varphi}_t(x_t)]$ and $\tilde{V}_t = \tilde{\Phi}_t^T\tilde{\Phi}_t + \lambda I_{m_t}$
\State Find the rows $u_1^T \ldots, u_{m_t}^T$ of $\tilde{V}_t^{-1/2}\tilde{\Phi}_t^T$
\State Set $\hat{r}_i = \sum_{\tau=1}^t u_{i,\tau}\bar{y}_{\tau}\mathbbm{1}_{|u_{i,\tau}\bar{y}_{\tau}| \le b_{\tau}}$ for $i \in [m_t]$
\State Set $\tilde{\theta}_t =\tilde{V}_t^{-1/2}[\hat{r}_1,\ldots,\hat{r}_{m_t}]^T$ 
\State Set $\tilde{\mu}_t(x) = \tilde{\varphi}_t(x)^T\tilde{\theta}_t$
\State Set $\tilde{\sigma}_t^2(x) = k(x,x)-\tilde{\varphi}_t(x)^T\tilde{\varphi}_t(x) +\lambda\tilde{\varphi}_t(x)^T\tilde{V}_t^{-1}\tilde{\varphi}_t(x)$
\EndFor
\end{algorithmic}
\end{algorithm}

Further, by adapting the regret analysis of ATA-GP-UCB in~\cite{chowdhury2019bayesian}, we have the following theorem for the regret upper bound of LDP-ATA-GP-UCB.
\begin{theorem}
\label{thm:ATA}
	Let $f \in \mathcal{H}_k(\mathcal{D})$ with $\norm{f}_{\mathcal{H}}\le B$ for all $x \in \mathcal{D}$ and noise $\eta_t$ is bounded by $R$. Fix $\epsilon > 0$,  $\varepsilon \in (0,1)$ and set $\rho = \frac{1+\varepsilon}{1-\varepsilon}$, and $v = B^2+R^2+8(B+R)^2/\epsilon^2$. Then, for any $\delta \in (0,1]$, LDP-ATA-GP-UCB with parameters $q = 6\rho\ln(4T/\delta)/\varepsilon^2$, $b_t = \sqrt{v/\ln(4m_tT/\delta)}$ and $\beta_{t+1} = B(1+\frac{1}{\sqrt{1-\varepsilon}}) + 4\sqrt{\ln(4m_tT/\delta) v m_t/\lambda}$, with probability at least $1-\delta$, has regret bound
	\begin{linenomath}
		\begin{align*}
		R_T = O\left(\hat{\rho} B\sqrt{T\gamma_T} + \frac{\rho}{\varepsilon^2} \gamma_T\sqrt{T}\sqrt{v\ln(T/\delta) \ln(\frac{\gamma_T T\ln(T/\delta)}{\delta}) } \right)
	\end{align*}
	\end{linenomath}
	where $\hat{\rho} = \rho\left(1+\frac{1}{\sqrt{1-\varepsilon}}\right)$.
\end{theorem}
\begin{remark}
	Note that by substituting the value of $v$ into the regret bound, we obtain that $R_T = \tilde{O}(\gamma_T\sqrt{T}/\epsilon)$. That is, it has a factor of $1/\epsilon$ compared to the non-private case, which matches the same scaling of $\epsilon$ in the lower bounds as shown in Theorem~\ref{thm:lb}. Moreover, LDP-ATA-GP-UCB enjoys the same scaling with respect to both $\gamma_T$ and $T$ as in the state-of-the-art non-private sub-Gaussian case. 
\end{remark}

Although the LDP-ATA-GP-UCB algorithm achieves almost optimal regret bound, it might be a `overkill' for the LDP setting. In particular, the original setting for the ATA-GP-UCB algorithm in~\cite{chowdhury2019bayesian} only assumes at most a finite variance. However, in our LDP setting, the corrupted reward $\bar{y}$ has all the moments being bounded and enjoys an exponential-type tail. In other words, although the additional Laplace noise causes the corrupted reward to be no longer sub-Gaussian, it still enjoys better properties compared to the general conditions for the ATA-GP-UCB algorithm to work. Therefore, it seems that there is some hope that we can design simple algorithm to achieve the same regret bound. Another issue of ATA-GP-UCB is its computational complexity. As pointed out by~\cite{shao2018almost} in the linear bandit setting (ATA-GP-UCB reduces to TOFU), for each round, it needs to truncate all the historical payoffs, which leads to a high complexity. 

Based on the discussions above, in the following, we will propose two novel algorithms that are also able to achieve almost optimal regret while substantially reducing the implementation and computational complexity of LDP-ATA-GP-UCB. 

\subsection{Raw Reward Truncation Algorithm and Regret}
In this section, instead of using the sophisticated truncation in the feature space as in LDP-ATA-GP-UCB, we turn to adopt the simple truncation on the raw rewards. In the general heavy-tail reward setting (with at most a finite variance),~\cite{chowdhury2019bayesian} proposed TGP-UCB algorithm which truncates the reward to zero if it is larger than a truncated point $b_t$ for round $t$. Specifically, for a finite variance case, the truncated point $b_t$ is $\theta(t^{1/4})$ in TGP-UCB, which finally leads to an regret bound of $\tilde{O}(T^{3/4})$. Hence, it has an additional factor $O(T^{1/4})$ when compared to the regret bound for the sub-Gaussian case. This means that we cannot directly adopt TGP-UCB to achieve the same regret bound as in LDP-ATA-GP-UCB of the last section.

However, as pointed before, the corrupted reward $\bar{y}$ has a nice exponential tail property. This suggests that a truncated point of order $O(\ln t)$ is enough, which will only in turn incurs an additional $O(\ln T)$ factor in the regret. Based on this idea, we propose the LDP-TGP-UCB algorithm, as described in Algorithm~\ref{alg:TGP}.

\begin{algorithm}[t]
\caption{LDP-TGP-UCB}\label{alg:TGP}
\begin{algorithmic}[1]
\State \textbf{Input:} Parameters $B$, $R$, $\epsilon >0$, $\lambda$, $\delta$.
\State \textbf{Set:} $K = B^2 + R^2 + 2\mathcal{L}^2$
        \For{{$t = 1,2,3,\ldots, T$}}
        	\State Set $b_{t-1} = B+R+\mathcal{L}\ln(t-1)$
        	\State Set $\beta_t = B + \frac{2\sqrt{2}}{\sqrt{\lambda}}b_{t-1}\sqrt{\gamma_{t-1} + \ln(1/\delta)}+\frac{1}{\sqrt{\lambda}}\sqrt{K(\ln(t-1)+1)}$
            \State Play $x_t = \arg\max_{x\in\mathcal{D}}\hat{\mu}_{t-1}(x) + \beta_t\sigma_{t-1}(x)$
            \State Receive private response $\bar{y}_t$ from CTL($\epsilon$).
            \State Set $\hat{y}_t = \bar{y}_t\mathbbm{1}_{|\bar{y}_t| \le b_t}$ and $\hat{Y}_t = [\hat{y}_1, \ldots, \hat{y}_t]^T$
            \State Set $\hat{\mu}_t(x) = k_t(x)^T(K_t + \lambda I)^{-1}\hat{Y}_t$
            \State Set $\sigma_t^2(x) = k(x,x)-k_t(x)^T(K_t + \lambda I)^{-1}k_t(x)$
		\EndFor
\end{algorithmic}
\end{algorithm}

Moreover, by refining the regret analysis of TGP-UCB in~\cite{chowdhury2019bayesian}, we can obtain the following theorem for the regret bound of LDP-TGP-UCB.
\begin{theorem}
\label{thm:TGP}
	Fix $\epsilon > 0$. Let $f \in \mathcal{H}_k(\mathcal{D})$ with $\norm{f}_{\mathcal{H}}\le B$ for all $x \in \mathcal{D}$ and the noise $\eta_t$ is bounded by $R$ for all $t$. Then, for any $\delta \in (0,1]$, LDP-TGP-UCB achieves, with probability at least $1-\delta$, the regret bound 
	\begin{linenomath}
		\begin{align*}
		R_T = O\left(\vartheta \sqrt{\ln T\gamma_T T} + \vartheta\sqrt{T}\ln T\sqrt{\gamma_T (\gamma_T + \ln(1/\delta))}\right),
	\end{align*}
	\end{linenomath}
	where $\vartheta = (B+R)/\epsilon$.
\end{theorem}
\begin{remark}
	As in LDP-ATA-GP-UCB, LDP-TGP-UCB is also able to achieve regret bound $\tilde{O}(\gamma_T\sqrt{T}/\epsilon)$. The advantage of LDP-TGP-UCB is its simple implementation in the sense that each reward is only trimmed once.
\end{remark}
\begin{proof}[Proof Sketch of Theorem~\ref{thm:TGP}] As in most GP-UCB like algorithms~\cite{chowdhury2017kernelized,chowdhury2019bayesian}, the key step boils down to establishing a (high-probability) confidence interval bound, i.e., in our setting, 
\begin{linenomath}
	\begin{align*}
	|f(x) - \hat{\mu}_t(x)| \le \beta_{t+1}\sigma_{t}(x).
\end{align*}
\end{linenomath}
To this end, with some linear algebra calculations, we have 
\begin{linenomath}
	\begin{align*}
	|f(x) - \hat{\mu}_t(x)| \le \left(B + \lambda^{-1/2} \norms{\sum_{\tau = 1}^t\hat{\eta}_{\tau}\varphi(x_{\tau})}_{V_t^{-1}} \right)\sigma_t(x),
\end{align*}
\end{linenomath}
where $\hat{\eta}_t = \hat{y}_t - f(x_t)$, $\varphi(x):=k(x,\cdot)$, which maps $x\in \mathbb{R}^d$ to RKHS $H$ associated with kernel function $k$ and $V_t = \Phi_t^T\Phi_t + \lambda I_{\mathcal{H}}$, $\Phi_t:=[\varphi(x_1)^T,\ldots, \varphi(x_t)^T]^T$.

The key term is $\norms{\sum_{\tau = 1}^t\hat{\eta}_{\tau}\varphi(x_{\tau})}_{V_t^{-1}}$, which can be handled by the self-normalized inequality if $\hat{\eta}_{\tau}$ is sub-Gaussian. However, in our setting, it is not. To overcome this issue, we will divide it into two parts. In particular, similar to~\cite{chowdhury2019bayesian}, we define $\xi_t = \hat{\eta}_t - \ex{\hat{\eta}_t \mid \mathcal{F}_{t-1}}$. Now, the key term can be written as
\begin{linenomath}
	\begin{align}
 	\norms{\sum_{\tau = 1}^t\hat{\eta}_{\tau}\varphi(x_{\tau})}_{V_t^{-1}}= \underbrace{\norms{\sum_{\tau = 1}^t {\xi}_{\tau}\phi(x_{\tau})}_{V_t^{-1}} }_{\mathcal{T}_1} + \underbrace{\norms{\sum_{\tau = 1}^t \ex{\hat{\eta}_{\tau} | \mathcal{F}_{\tau-1}}\phi(x_{\tau})}_{V_t^{-1}}.}_{\mathcal{T}_2}
 \end{align} 
\end{linenomath}
 For $\mathcal{T}_1$, note that $\xi_t = \hat{y}_t -\ex{\hat{y}_t \mid \mathcal{F}_{t-1}}$, which is bounded by $2b_t$, and hence sub-Gaussian. Thus, by the self-normalized inequality for the RKHS-valued process in~\cite{durand2018streaming,chowdhury2019bayesian}, we can bound $\mathcal{T}_1$ as follows
 \begin{linenomath}
 	\begin{align*}
 	\mathcal{T}_1 \le 2b_t\sqrt{2(\gamma_t + \ln(1/\delta))}.
 \end{align*}
 \end{linenomath}
 For $\mathcal{T}_2$, with some linear algebra, we can first bound it as $\sqrt{\sum_{\tau=1}^t \ex{\hat{\eta}_{\tau} | \mathcal{F}_{\tau-1}}^2}$. Further, note that $\ex{\hat{\eta}_{\tau} | \mathcal{F}_{\tau-1}} = -\ex{\bar{y}_{\tau}\mathbbm{1}_{|\tilde{y}_{\tau}| > b_{\tau}}|\mathcal{F}_{\tau-1}}$. Hence, by Cauchy-Schwartz inequality with $b_{\tau} = B+R+\mathcal{L}\ln \tau$, we have 
 \begin{linenomath}
 	\begin{align*}
 	\ex{\hat{\eta}_{\tau} | \mathcal{F}_{\tau-1}}^2 \le \ex{\bar{y}_{\tau}^2|\mathcal{F}_{\tau-1}} \mathbbm{P}(|L| > \mathcal{L}\ln \tau) \le K\frac{1}{\tau},
 \end{align*}
 \end{linenomath}
where $K:=B^2+R^2+2\mathcal{L}^2$. The last inequality holds since $|L| \sim \text{Exp}(1/\mathcal{L})$. Therefore, by the property of Harmonic sum, we have 
\begin{linenomath}
	\begin{align*}
	\mathcal{T}_2 \le\sqrt{K(\ln t + 1)}.
\end{align*}
\end{linenomath}
Hence, the (high-probability) confidence interval bound is obtained by setting
\begin{linenomath}
	\begin{align*}
	\beta_{t+1} = B + \frac{2\sqrt{2}}{\sqrt{\lambda}}b_{t}\sqrt{\gamma_{t} + \ln(1/\delta)}+\frac{1}{\sqrt{\lambda}}\sqrt{K(\ln t+1)}.
\end{align*}
\end{linenomath}
The full proof is relegated to Appendix.
\end{proof}

It is worth pointing out the truncation trick used in LDP-TGP-UCB also sheds light on the regret bounds for non-private BO under payoffs that are beyond sub-Gaussian, e.g., sub-Weibull which includes sub-Gaussian and sub-exponential as special cases~\cite{vladimirova2019sub}. More specifically, according to~\cite{vladimirova2019sub}, a random variable $X$ is said to be a sub-Weibull with tail parameter $\theta$, i.e., $X\sim \text{subW}(\theta)$, if for some constants $a$ and $b$ such that
\begin{linenomath}
	\begin{align}
\label{eq:subWeibull}
	\mathbb{P}(|X| \ge x) \le a \exp(-bx^{1/\theta}),\quad\text{for all } x > 0.
\end{align}
\end{linenomath}
It can be seen that sub-Gaussian and sub-exponential distributions are special cases of sub-Weibull with $\theta = 1/2$ and $\theta = 1$, respectively.
Thus, instead of choosing the truncation point $b_t = O(\ln t)$ as in LDP-TGP-UCB, one turn to choose $b_t = O((\ln t)^{\theta})$, which in turn only incurs a log factor in the regret bound. As a result, with this simple truncation, the non-private BO under sub-Weibull noise is still $\tilde{O}(\gamma_T\sqrt{T})$.


\subsection{Median of Means Approximate (MoMA) Algorithm and Regret}
In this section, we will introduce a new BO method that is able to achieve almost optimal regret bounds under general heavy-tailed payoffs. Hence, the LDP setting is just a special case. This new method is mainly inspired by the MENU algorithm in~\cite{shao2018almost}, which is introduced to handle the heavy-tailed payoffs in the linear bandit setting. Moreover, it has been shown to have a lower complexity than TOFU (which is the core of ATA-GP-UCB). The key idea in MENU is based on median of means techniques~\cite{bubeck2013bandits}. The main challenge in generalizing MENU to the BO setting is to handle the possibly infinite feature dimension associated with the kernel function. To this end, we will again use kernel approximation techniques (i.e., Nystr\"{o}m approximation) developed in~\cite{calandriello2019gaussian,chowdhury2019bayesian}. However, instead of updating the approximations after every iteration as in ATA-GP-UCB, in our new method the approximation is only updated after each `epoch', which is composed of multiple iterations. This further reduces its complexity.

\begin{figure}[t]
\centering
\includegraphics[width=2.5in]{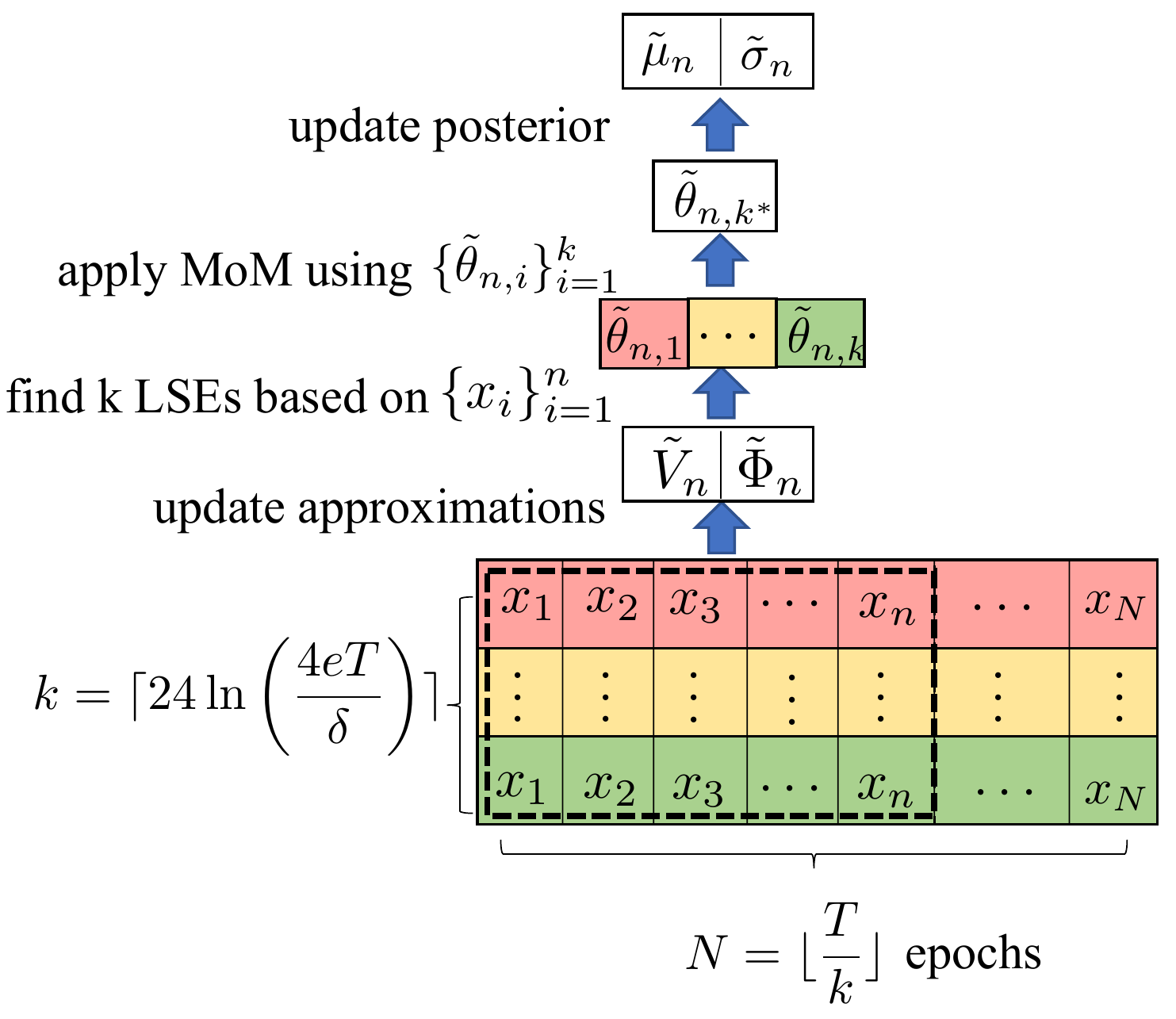} 
\caption{Illustration of MoMA-GP-UCB}
\label{fig2}
\end{figure}

This new algorithm is called MoMA-GP-UCB, which is presented in Algorithm~\ref{alg:MoMA-GP-UCB}. With the aid of Fig.~\ref{fig2} (adapted from~\cite{shao2018almost}), we can easily see that the total number of $T$ iterations are divided into $N$ epochs, each consisted of $k$ iterations. The algorithm will loop over each $n=1,\ldots, N$ epoch. Within each epoch $n$, a point $x_n$ is selected in a GP-UCB fashion, and the selected point $x_n$ will be played $k$ times with corresponding rewards. Then, the kernel approximation terms are updated, i.e., $\tilde{\varphi}_n$,$\tilde{\Phi}_n$ and $\tilde{V}_n$. Following this update, it will calculate $k$ least-square-estimates (LSE), each is based on the rewards along each row $j\in [k]$ (e.g., using the data in the pink row to generate the pink LSE, and similarly green data for the green LSE). Next, it applies median-of-means techniques to find the best LSE $\tilde{\theta}_{n,k^*}$ for epoch $n$. Finally, the posterior mean and variance are updated.

Now, we have the following theorem for the regret bound of MoMA-GP-UCB under general heavy-tailed payoffs.
\begin{theorem}
\label{thm:MoMA}
	Let $f \in \mathcal{H}_k(\mathcal{D})$ with $\norm{f}_{\mathcal{H}}\le B$ for all $x \in \mathcal{D}$. Assume that $\ex{|\eta_t|^{1+\alpha} \mid \mathcal{F}_{t-1}} \le c $. Fix $\varepsilon \in (0,1)$ and set $\rho = \frac{1+\varepsilon}{1-\varepsilon}$. Then, for any $\delta \in (0,1]$, MoMA-GP-UCB with parameters $q = 6\rho\ln(4T/\delta)/\varepsilon^2$, $k = \lceil 24\ln\left(\frac{4eT}{\delta}\right)\rceil$, and $\beta_{n+1} = B(1+\frac{1}{\sqrt{1-\varepsilon}}) + 3\left((9m_nc)^{\frac{1}{1+\alpha}}n^{\frac{1-\alpha}{2(1+\alpha)}}\right)$, with probability at least $1-\delta$, has regret bound
	\begin{linenomath}
		\begin{align*}
		R_T = O\left(\hat{B}\sqrt{\gamma_T T \ln \frac{T}{\delta}} + Z \ln \frac{T}{\delta}c^{\frac{1}{1+\alpha}}T^{\frac{1}{1+\alpha}}\gamma_T^{\frac{3+\alpha}{2(1+\alpha)}}\right),
	\end{align*}
	\end{linenomath}
	where $\hat{B}=\rho B(1+\frac{1}{\sqrt{1-\varepsilon}})$ and $Z = \left(\frac{\rho^{3+\alpha}}{\varepsilon^2}\right)^{\frac{1}{1+\alpha}}$.
\end{theorem}
\begin{remark}
	Note that when $\alpha = 1$, MoMA-GP-UCB recovers the same regret bound $\tilde{O}(\gamma_T \sqrt{T})$ as in the sub-Gaussian case. Moreover, for the special linear kernel case, substituting $\gamma_T = O(d\ln T)$, the bound in Theorem~\ref{thm:MoMA} recovers the regret bound in~\cite{shao2018almost} up to logarithmic factor.
\end{remark}

\begin{proof}[Proof Sketch of Theorem~\ref{thm:MoMA}]
	The proof is mainly inspired by~\cite{shao2018almost,chowdhury2019bayesian}. The key step is again a (high-probability) confidence interval bound, i.e., 
	\begin{linenomath}
		\begin{align}
	\label{eq:CI}
	|f(x) - \tilde{\mu}_n(x)| \le \beta_{n+1}\tilde{\sigma}_{n}(x).
	\end{align}
	\end{linenomath}
	Assume that the kernel approximation is $\varepsilon$-accurate, the LHS of Eq.~\eqref{eq:CI} can be bounded by 
	\begin{linenomath}
		\begin{align}
		|f(x) - \tilde{\mu}_n(x)| &\le B(1+\frac{1}{\sqrt{1-\varepsilon}})\tilde{\sigma}_{n}(x)\nonumber\\
		+& \lambda^{-1/2}\norms{\tilde{V}_n^{-1}\tilde{\Phi}_n^Tf_n - \tilde{\theta}_{n,k^*} }_{\tilde{V}_n} \tilde{\sigma}_{n}(x)\label{eq:theta},
	\end{align}
	\end{linenomath}
	where $f_n = [f(x_1),\ldots,f(x_n)]^T$, i.e., a vector containing $f$'s evaluations up to epoch $n$.
	Now, we need to focus on the term $\norms{\tilde{V}_n^{-1}\tilde{\Phi}_n^Tf_n - \tilde{\theta}_{n,k^*} }_{\tilde{V}_n}$. To this end, we first establish the following result regarding the $k$ LSEs. In particular, for $j \in[k]$, we have 
	\begin{linenomath}
		\begin{align*}
		\mathbb{P}\left( \norms{\tilde{V}_n^{-1}\tilde{\Phi}_n^Tf_n - \tilde{\theta}_{n,j} }_{\tilde{V}_n} \le  \gamma\right) \ge \frac{3}{4},
	\end{align*}
	\end{linenomath}
	where $\gamma := (9m_n c)^{\frac{1}{1+\alpha}}n^{\frac{1-\alpha}{2(1+\alpha)} }$. Based on this result, by the choice of $k^*$, we obtain that if $k = \lceil 24\ln\left(\frac{eT}{\delta}\right)\rceil$, then for all $n \in [N]$, with probability at least $1-\delta$,
	\begin{linenomath}
		\begin{align}
	\label{eq:moma-ub}
		\norms{\tilde{V}_n^{-1}\tilde{\Phi}_n^Tf_n - \tilde{\theta}_{n,k^*} }_{\tilde{V}_n} \le 3\gamma.
	\end{align}
	\end{linenomath}
	Combining Eqs.~\eqref{eq:moma-ub} and~\eqref{eq:theta}, yields that, under the event that the kernel approximation is $\varepsilon$-accurate, with probability at least $1-\delta$, 
	\begin{linenomath}
		\begin{align*}
		|f(x) - \tilde{\mu}_n(x)| &\le \left(B(1+\frac{1}{\sqrt{1-\varepsilon}})+\lambda^{-1/2}3\gamma \right) \tilde{\sigma}_{n}(x)
	\end{align*}
	\end{linenomath}
	for all $n \in [N]$ when $k = \lceil 24\ln\left(\frac{eT}{\delta}\right)\rceil$. Since for any $\delta$, the kernel approximation (under given parameters) is $\varepsilon$-accurate with probability at least $1-\delta$, by the virtue of union bound, we have that when $k = \lceil 24\ln\left(\frac{2eT}{\delta}\right)\rceil$, 
	\begin{linenomath}
		\begin{align}
		|f(x) - \tilde{\mu}_n(x)| &\le \beta_{n+1} \tilde{\sigma}_{n}(x)
	\end{align}
	\end{linenomath}
	for all $n \in [N]$, where $\beta_{n+1}:=B(1+\frac{1}{\sqrt{1-\varepsilon}})+\lambda^{-1/2}3\gamma$.
	Finally, by the nice properties of Nystr\"{o}m approximation, we can obtain that with probability at least $1-\delta$, both $m_n = O(\frac{\rho^2}{\varepsilon^2}\gamma_n\ln(T/\delta) )$ and $\tilde{\sigma}_{n-1}(x_n) \le \rho {\sigma}_{n-1}(x_n)$. Then, the regret bound follows from that $R_T = 2k \sum_{n=1}^N \beta_n \tilde{\sigma}_{n-1}(x)$ along with standard GP-UCB analysis.  The full proof is relegated to Appendix.
\end{proof}

\begin{algorithm}[t]
\caption{MoMA-GP-UCB}\label{alg:MoMA-GP-UCB}
\begin{algorithmic}[1]
\State \textbf{Input:} Parameters $\lambda$, $\delta$, $\{\beta_t\}_{t\ge1}$, and $q$.
\State \textbf{Set:} $\tilde{\mu}_0(x) = 0$ and $\tilde{\sigma}_0(x) = k(x,x)$ for all $x \in \mathcal{D}$.
\State \textbf{Set:} $k = \lceil 24\ln\left(\frac{4eT}{\delta}\right)\rceil$ and $N = \lfloor \frac{T}{k} \rfloor$.
\For{$n = 1,2,3,\ldots, N$}
\State $x_n = \argmax_{x\in\mathcal{D}}\tilde{\mu}_{n-1}(x) + \beta_{n}(x)\tilde{\sigma}_{n-1}(x)$
\State Play $x_n$ with $k$ times and observe rewards $y_{n,1}, y_{n,2},\ldots, y_{n,k}$. 
\State Set $\tilde{\varphi}_n(x) = \text{Nystr\"{o}mEmbedding}(\{(x_i,\tilde{\sigma}_{n-1}(x_i))\}_{i=1}^n,q)$ 
\State Set $m_n$ as the dimension of $\tilde{\varphi}_n$
\State Set $\tilde{\Phi}_n^T = [\tilde{\varphi}_n(x_1), \ldots, \tilde{\varphi}_n(x_n)]$ and $\tilde{V}_n = \tilde{\Phi}_n^T\tilde{\Phi}_n + \lambda I_{m_n}$
\State For $j\in[k]$, $\tilde{\theta}_{n,j} = \tilde{V}_n^{-1}\sum_{i=1}^n y_{i,j}\tilde{\varphi}_n(x_i)$
\State For $j\in[k]$, $r_j = $ median$(\{ \norms{ \tilde{\theta}_{n,j} - \tilde{\theta}_{n,s}  }_{\tilde{V}_n}: s \in [k]\setminus j  \})$
\State Set $k^* =\arg\min_{j\in[k]}r_j$
\State Set $\tilde{\mu}_n(x) = \tilde{\varphi}_n(x)^T\tilde{\theta}_{n,k^*}$
\State Set $\tilde{\sigma}_n^2(x) = k(x,x)-\tilde{\varphi}_n(x)^T\tilde{\varphi}_n(x) +\lambda\tilde{\varphi}_n(x)^T\tilde{V}_n^{-1}\tilde{\varphi}_n(x)$
\EndFor
\end{algorithmic}
\end{algorithm}


Now, we can easily design the private version of it, called LDP-MoMA-GP-UCB. The only difference is line $6$, in which LDP-MoMA-GP-UCB received private response from CTL($\epsilon$). Thus, we can directly obtain the regret bound of LDP-MoMA-GP-UCB as a corollary of Theorem~\ref{thm:MoMA}.
\begin{corollary}
	Let $f \in \mathcal{H}_k(\mathcal{D})$ with $\norm{f}_{\mathcal{H}}\le B$ for all $x \in \mathcal{D}$ and noise $\eta_t$ is bounded by $R$. Fix $\epsilon > 0$,  $\varepsilon \in (0,1)$ and set $\rho = \frac{1+\varepsilon}{1-\varepsilon}$, and $c = R^2+8(B+R)^2/\epsilon^2$. Then, for any $\delta \in (0,1]$, LDP-MoMA-GP-UCB with parameters $q = 6\rho\ln(4T/\delta)/\varepsilon^2$, $k = \lceil 24\ln\left(\frac{4eT}{\delta}\right)\rceil$ and $\beta_{n+1} = B(1+\frac{1}{\sqrt{1-\varepsilon}}) + 3\left((9m_nc)^{\frac{1}{1+\alpha}}n^{\frac{1-\alpha}{2(1+\alpha)}}\right)$, with probability at least $1-\delta$, has regret bound
	\begin{linenomath}
		\begin{align*}
		R_T = O\left(\hat{B}\sqrt{\gamma_T T \ln \frac{T}{\delta}} + Z \ln \frac{T}{\delta} \frac{B+R}{\epsilon} \gamma_T\sqrt{T}\right),
	\end{align*}
	\end{linenomath}
	where $\hat{B}=\rho B(1+\frac{1}{\sqrt{1-\varepsilon}})$ and $Z = \left(\frac{\rho^{3+\alpha}}{\varepsilon^2}\right)^{\frac{1}{1+\alpha}}$.
\end{corollary}

\subsection{Unbounded Noise Case}
In previous sections, we assume that $\eta_t$ is bounded by $R$. Hence, by Laplace mechanism, we can achieve $\epsilon$-LDP. In the case of unbounded noise, we show that Laplace mechanism can ensure $(\epsilon,\delta)$-LDP (weaker than $\epsilon$-LDP) while still achieving almost optimal regret bounds for a large class of noise. More specifically, consider the case $\eta_t$ is $i.i.d$ and $\eta_t \sim \text{subW}(\theta)$ as defined in Eq.~\eqref{eq:subWeibull}. Then, it is easy to see that there exists a constant $K_1$ such that $\mathbb{P}(\exists t\in[T], |\eta_T| > K_1 \ln(T/\delta)^{\theta} ) \le \delta$. Thus, by letting $R = K_1 \ln(T/\delta)^{\theta}$ in the Laplace mechanism (i.e., CTL($\epsilon$)), the previous three private algorithms can protect $(\epsilon,\delta)$-LDP while only incurring an additional logarithmic factor in the regret bounds.

%% file: simulations.tex
\section{Experiments}

We conduct experiments to compare the performance of the three private algorithms (i.e., LDP-ATA-GP-UCB, LDP-TGP-UCB, LDP-MoMA-GP-UCB) and the performance of three non-private BO methods for general heavy-tailed payoffs (i.e., ATA-GP-UCB, TGP-UCB in~\cite{chowdhury2019bayesian} and MoMA-GP-UCB proposed in this paper). As in~\cite{chowdhury2019bayesian}, the parameters used for each algorithm are set order-wise similar to those recommended by the theorems. We run each algorithm for $10$ independent trials and plot the average of cumulative regret along with time evolution. 

\subsection{Datasets and Settings}
\textbf{Synthetic data.} The domain $\mathcal{D}$ is generated by discretizing $[0,1]$ uniformly into $100$ points. The black-box function $f = \sum_{i=1}^p a_i k(\cdot,x_i)$ is generated by uniformly sampling $a_i \in [-1,1]$ and support points $x_i \in \mathcal{D}$ with $p = 100$. The parameters for the kernel function are $l = 0.2$ for $k_{\text{SE} }$ and $l = 0.2$, $\nu = 2.5$ for $k_{\text{Mat\'ern} }$. We set $B = \max_{x \in \mathcal{D}}|f(x)|$ and $y(x) = f(x) + \eta$. For the LDP case, the noise $\eta$ is uniformly sampled in $[-1,1]$ and hence $R = 1$. For the non-private heavy-tailed case, the noise $\eta$ are samples from the Student's $t$-distribution with $3$ degrees of freedom. Hence, $v = B^2 + 3$ and $c = 3$.

\textbf{Light sensor data.} This data is collected in the CMU Intelligent Workplace in Nov 2005,
which is available online as Matlab structure\footnote{\url{http: //www.cs.cmu.edu/~guestrin/Class/10708-F08/projects}} and contains locations of 41 sensors, 601 train samples and 192 test samples. We use it in the context of finding the maximum average reading of the sensors. For fair comparison, the settings for this dataset follow from~\cite{chowdhury2019bayesian},which has shown that the payoffs are heavy-tailed. In particular, $f$ is set as empirical average of the test samples, with $B$ set as its maximum, and $k$ is set as the empirical covariance of the normalized train samples. 
The noise is estimated by taking the difference between the test samples and its empirical mean (i.e., $f$), and $R$ is set as the maximum. Here, we consider $\alpha =1$, set $v$ as the empirical mean of the squared readings of test samples, and $c$ is the empirical mean of the squared noise.

\textbf{Stock market data.} This dataset is the adjusted closing price of $29$ stocks from January 4th, 2016 to April 10th, 2019. We use it in the context of identifying the most profitable stock in a given pool of stocks. As verified in~\cite{chowdhury2019bayesian}, the rewards follows from heavy-tailed distribution. We take the empirical mean of stock prices as our objective function $f$ and empirical covariance of the normalized stock prices as our kernel function $k$. The noise is estimated by taking the difference between the raw prices and its empirical mean (i.e., $f$), with $R$ set as the maximum.
Consider $\alpha = 1$, with $v$ set as the empirical mean of the squared prices and $c$ set as the empirical mean of squared noise.

\subsection{Results}
\begin{figure*}[t]\centering
		\begin{subfigure}[b]{0.32\textwidth}
			\includegraphics[scale=0.14]{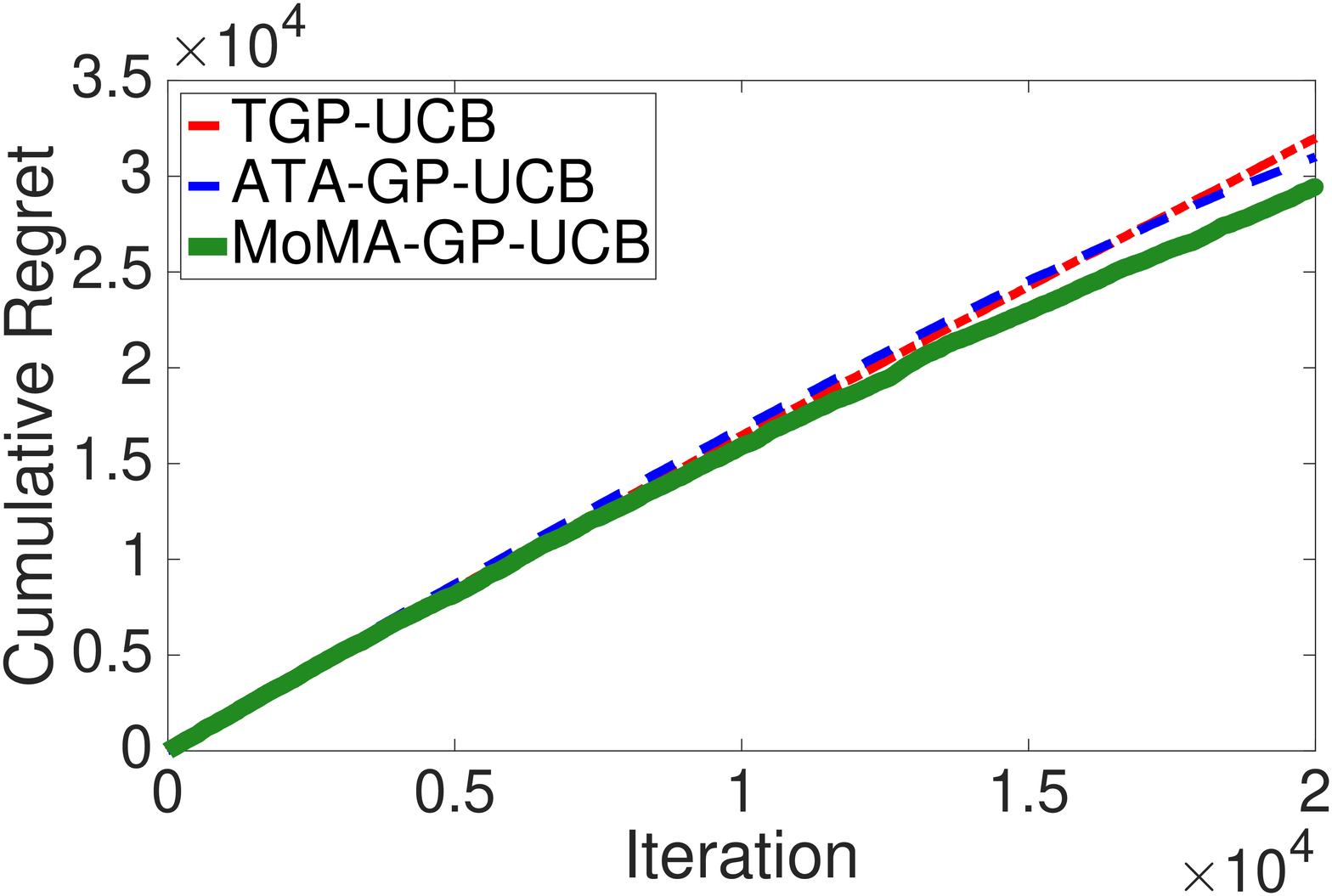}
			\caption{LDP, $\epsilon = 1$, Synthetic data, $k_{\text{Mat\'ern} }$}
		\end{subfigure}\ \ 
		\begin{subfigure}[b]{0.32\textwidth}
			\includegraphics[scale=0.14]{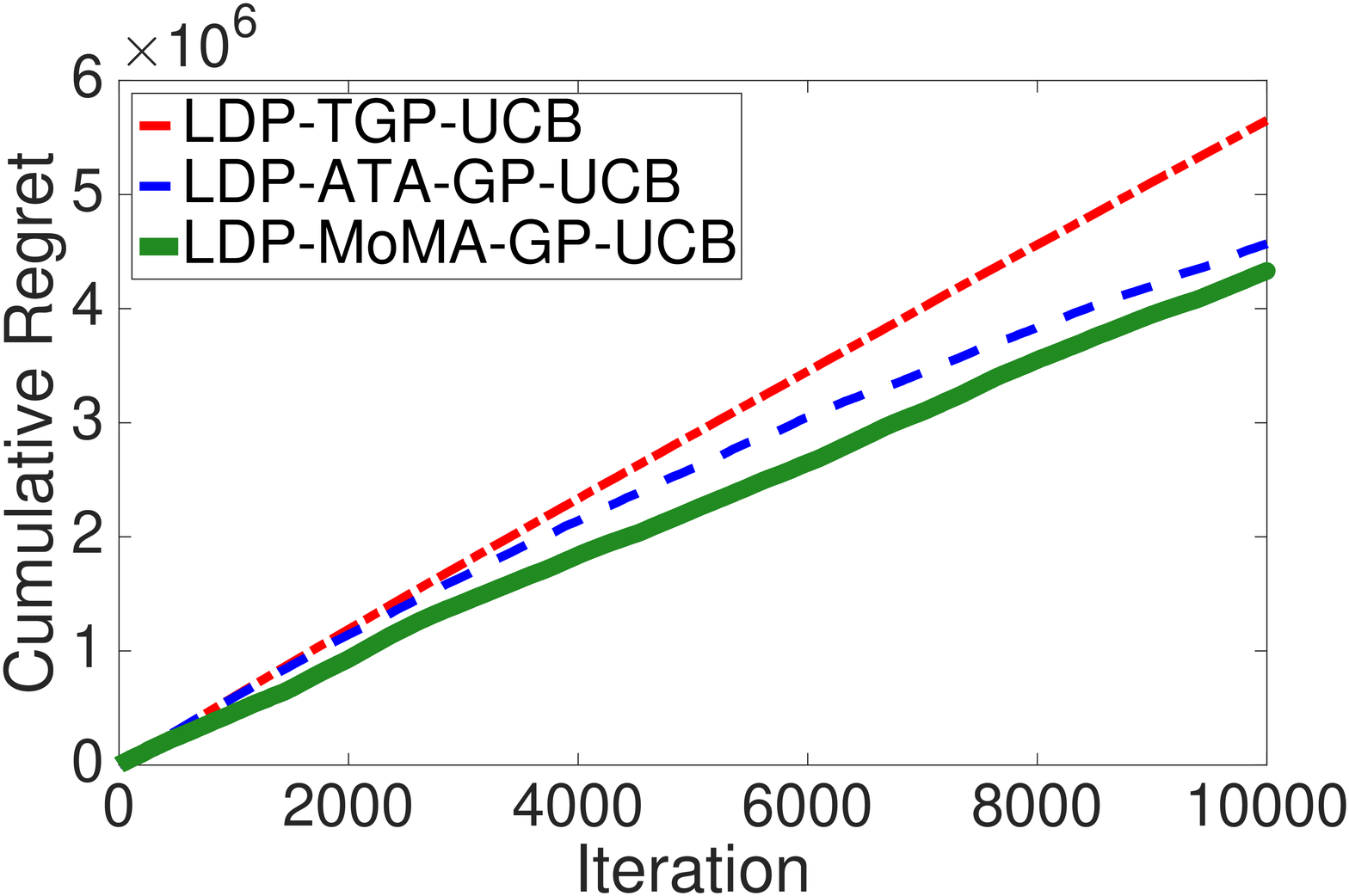}
			\caption{LDP, $\epsilon = 0.5$, Light sensor data}
		\end{subfigure}\ \ 
		\begin{subfigure}[b]{0.32\textwidth}
			\includegraphics[scale=0.14]{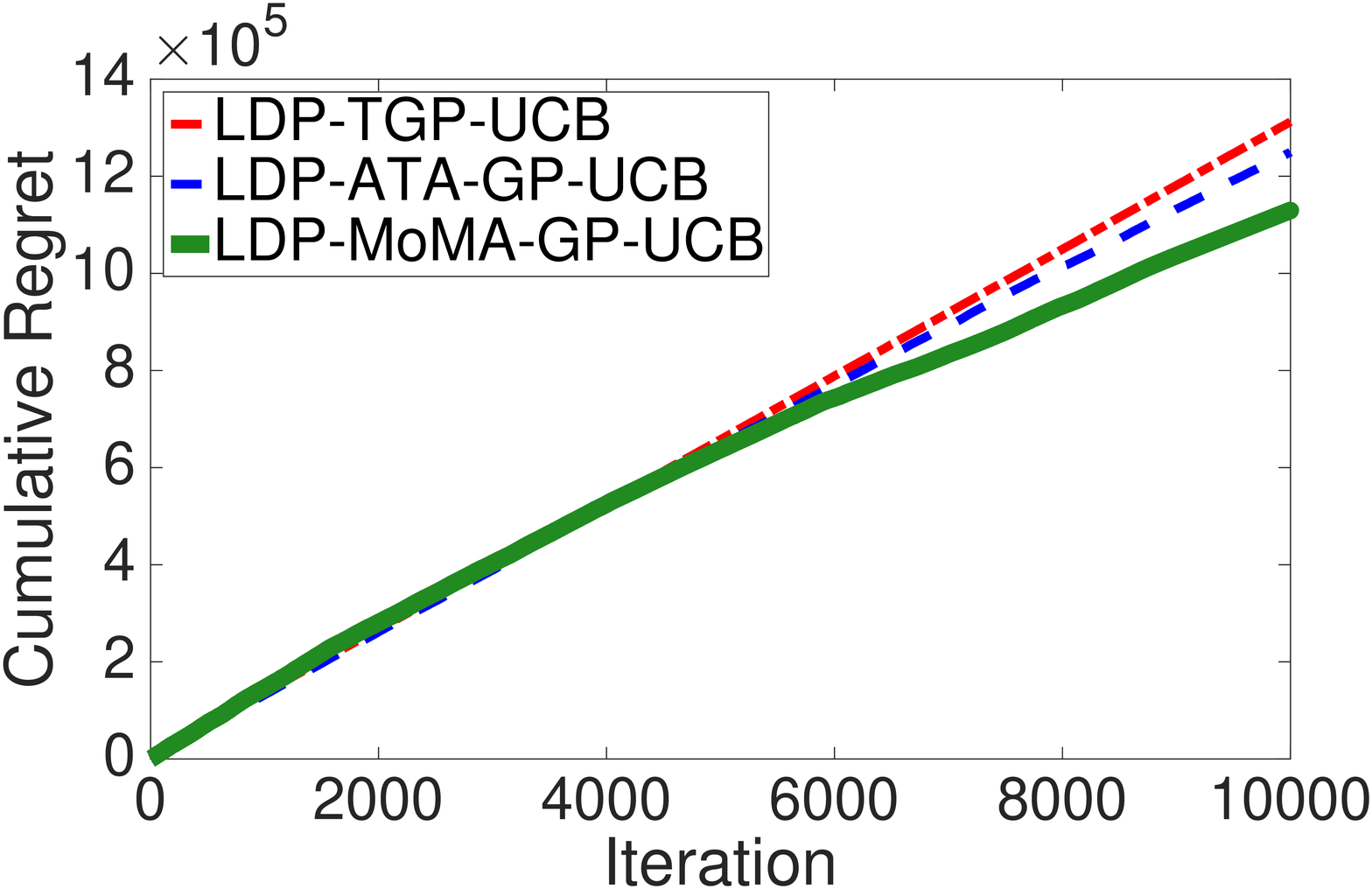}
			\caption{LDP, $\epsilon = 1$, Stock market data.}
		\end{subfigure}\\
		\begin{subfigure}[b]{0.32\textwidth}
			\includegraphics[scale=0.14]{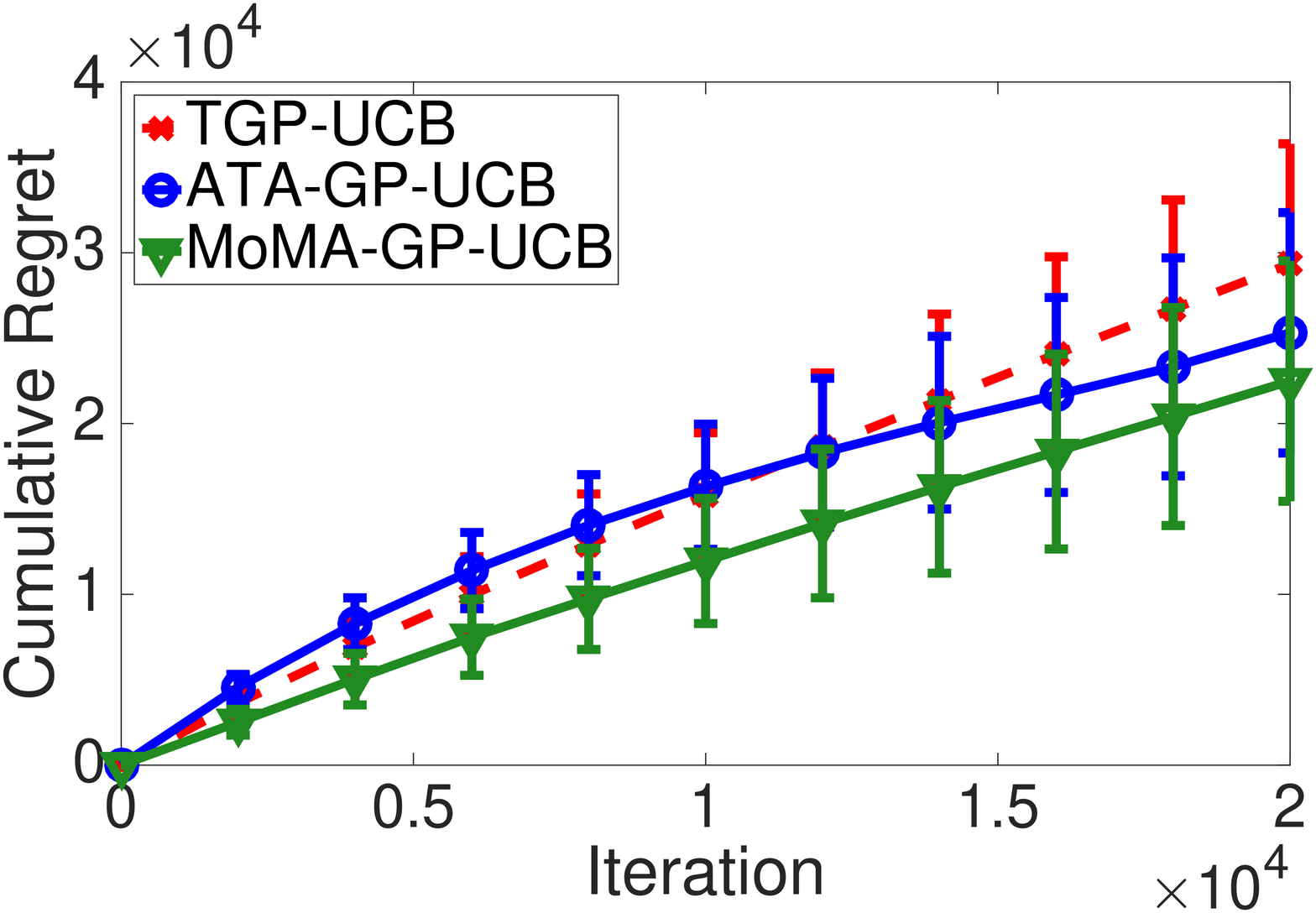}
			\caption{Non-private, Synthetic data, $k_{\text{SE} }$}
		\end{subfigure}\ \ 
		\begin{subfigure}[b]{0.32\textwidth}
			\includegraphics[scale=0.14]{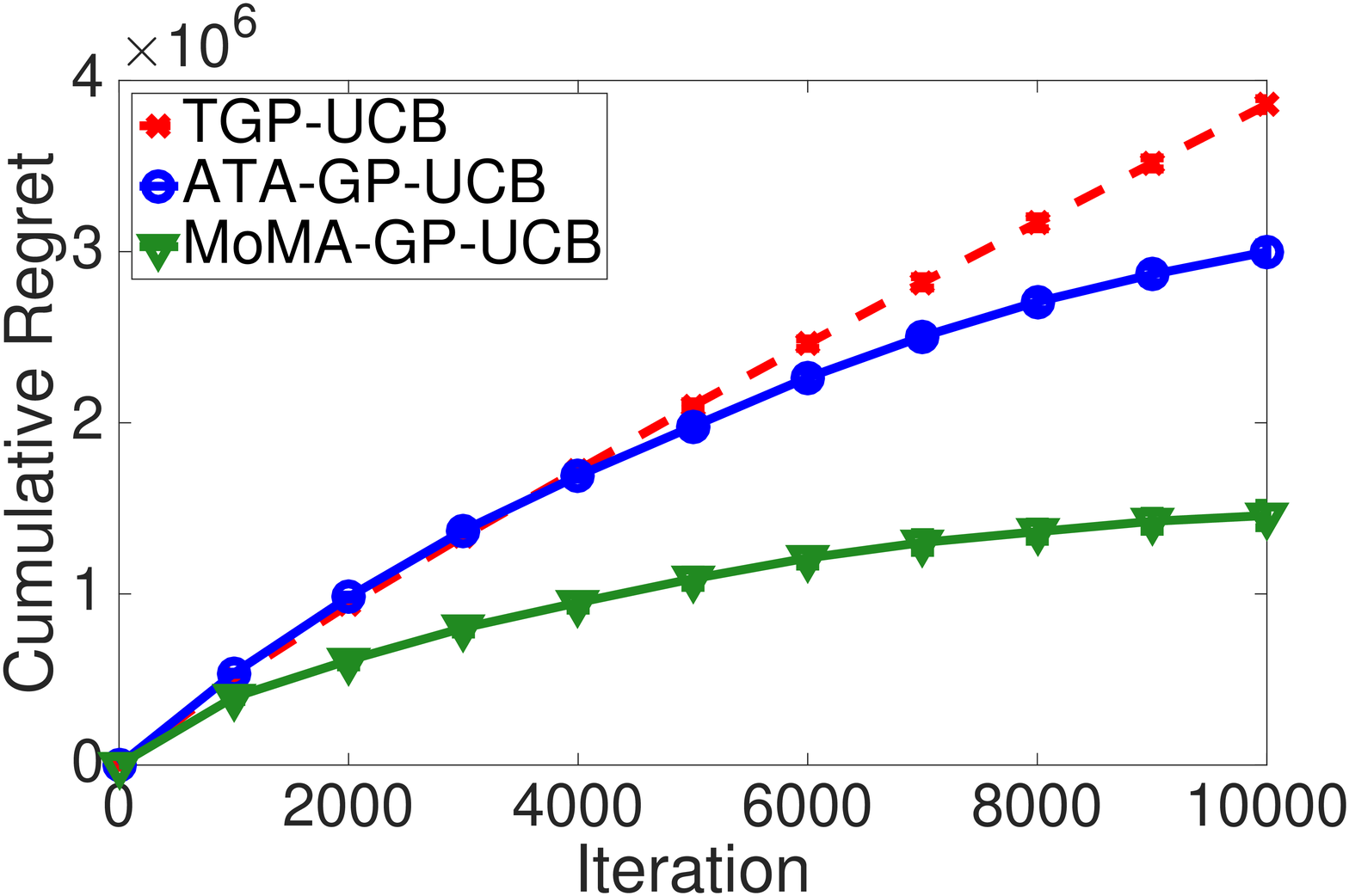}
			\caption{Non-private, Light sensor data}
		\end{subfigure}\ \ 
		\begin{subfigure}[b]{0.32\textwidth}
			\includegraphics[scale=0.14]{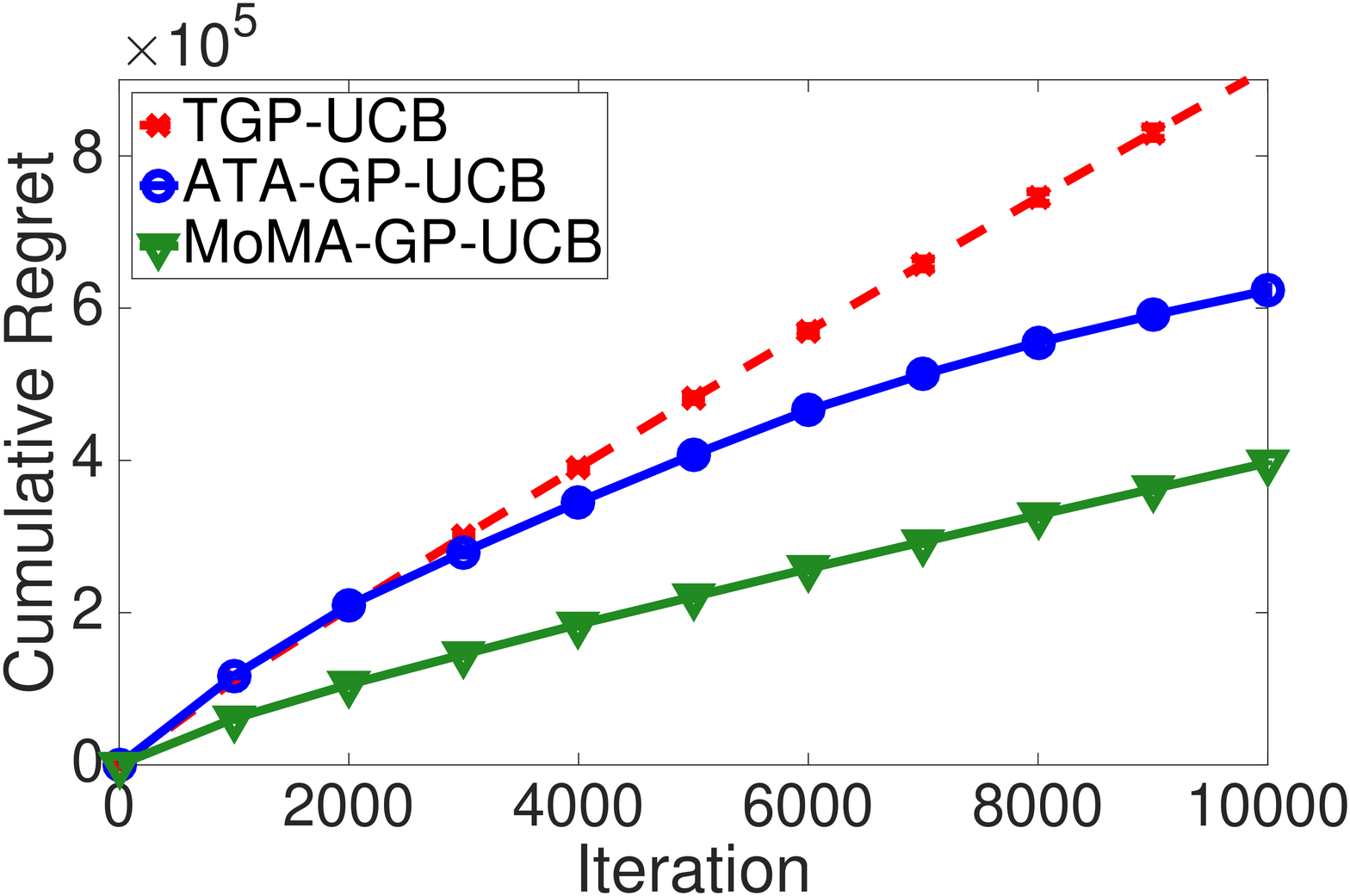}
			\caption{Non-private, Stock market data}
		\end{subfigure}
		\caption{(a)-(c) Cumulative regrets for three LDP algorithms; (d)-(f) Cumulative regrets (and standard variance) for non-private versions on heavy-tailed data.}\label{fig:regret}
\end{figure*}

From Figure~\ref{fig:regret}, we can see that MoMA-GP-UCB (or LDP-MoMA-GP-UCB) tends to empirically outperform the other two algorithms in both non-private and private settings. We also conduct additional experiments (relegated to Appendix), and similar observations are obtained. Note that similar to~\cite{chowdhury2019bayesian}, the high error bar in (d) is because a different $f$ is chosen for each trial.

%% file: appendix.tex
\newpage
\onecolumn
\section{Proof of Theorem~\ref{thm:lb}}
In this section, we will provide the detailed proof for the lower bounds in Theorem~\ref{thm:lb}. As pointed out earlier, our analysis build on the lower bounds for non-private BO in~\cite{chowdhury2019bayesian,scarlett2017lower}, but with important differences to handle the requirement of $\epsilon$-LDP under \emph{any} private mechanism. Roughly speaking, we follow the same construction of a function class in~\cite{scarlett2017lower} (with some slight changes as noted in footnotes) and the reward construction in~\cite{chowdhury2019bayesian} for heavy-tailed payoffs. To handle the LDP requirement, we take the strategy of relating a non-private version KL divergence to a private version KL divergence. It is worth noting that recently high probability (rather than expected) regret lower bounds in the case of non-private BO with Gaussian noise have also been derived~\cite{cai2020lower}. By using the same method of handling LDP in this proof, one can directly extend the results in~\cite{cai2020lower} to establish private high probability lower bounds.

In the following, to make our proof self-contained, we first briefly review the necessary backgrounds and notations used in previous papers~\cite{chowdhury2019bayesian,scarlett2017lower,cai2020lower} for establishing the lower bounds.
\subsection{Construction of a function class}
\label{sec:construction}
The high-level idea of this process is to construct a set of `difficult' functions, i.e., $\mathcal{F} = \{f_1,\ldots,f_M\}$, by shifting a common function $g$ by different amount. By `difficult', we mean that this function class satisfies that any single point $x$ can only be $\Delta$-optimal for at most one function in this set. Then, we take $f$ as a uniformly sampled function from $\mathcal{F}$. The idea is that if we can lower bound the regret averaged over the set, then there must exist a particular value of $m$ such that the same
lower bound applies to $f_m$. The details are as follows.
\begin{itemize}
	\item The common function $g$ is a function on $\mathbb{R}^d$ with the following properties:
	\begin{enumerate}
		\item $g$ is bounded in the RKHS norm, i.e., $\norm{g}_{\mh} \le B$.
		\item $g$ takes values in $[-2\Delta, 2\Delta]$ with a peak value of $2\Delta$ at $x = 0$, for some $\Delta>0$ determined later.
		\item $g$ takes values strictly less than $\Delta$, i.e., $g(x) < \Delta$, whenever $\norm{x}_{\infty} > \frac{w}{2}$, for some $w$ to be chosen later\footnote{Note that, in all previous papers~\cite{chowdhury2019bayesian,scarlett2017lower,cai2020lower}, the condition is $\norm{x}_{\infty} > {w}$ for property 3. However, it turns out to be insufficient to construct the `difficult' function class, i.e., any single point $x$ can only be $\Delta$-optimal for at most one function among $M$ functions constructed by the shifting process.}.  
	\end{enumerate}
	\item The shifting process of $g$ on the domain $\mathcal{D} = [0,1]^d$ is as follows. First, by taking a step size of $w$ in each dimension, we can construct a grid of size $M = \lfloor\left(\frac{1}{w}\right)^d \rfloor$. Then, each function $f_j$ is constructed by centering $g$ on the $j$th point of the grid, and then cropping to $\mathcal{D} = [0,1]^d$. Hence, we obtain $M$ functions, i.e., $\mathcal{F} = \{f_1,\ldots,f_M\}$. Moreover, by the properties of $g$ mentioned above, we ensure that any $\Delta$-optimal point for $f_i$ fails to be a $\Delta$-optimal point for any other function $f_j$ where $j \neq i$.
\end{itemize}

Now, it only remains to choose $g$, $\Delta$, and $w$ (and hence $M$).
\begin{itemize}
	\item Choose $g(x) = \frac{2\Delta}{h(0)}h\left(\frac{2x\zeta}{w}\right)$, for some absolute constant $\zeta > 0$, where $h$ is the inverse Fourier transform of the multi-dimensional bump function: $H(z) = e^{-\frac{1}{1-\norm{z}_2^2}}\mathbbm{1}\{\norm{z}_2^2 \le 1\}$. Since $H$ is real and symmetric, we have the maximum of $h$ is attained at $x = 0$, and hence $g$ has its maximum at $x = 0$ with value $2\Delta$. Moreover, since $H$ has finite energy, $h(x) \to 0$ when $\norm{x}_2 \to \infty$. Then, there exits a constant $\zeta >0$ such that $h(x) < \frac{1}{2}h(0)$ when $\norm{x}_{\infty} > \zeta$. Thus, when $\norm{x}_{\infty} > \frac{w}{2}$, $g(x) < \Delta$, as required before\footnote{Note that, we choose $\frac{2x\zeta}{w}$ in $h$ rather than $\frac{x\zeta}{w}$ as in previous papers, since we actually require that $g(x) < \Delta$ for any $\norm{x}_{\infty} > \frac{w}{2}$. Thus, all $\zeta$ in previous papers should be replaced by $2\zeta$. However, it is just a constant difference, and does not affect the final results in previous papers.}.
	\item Now, we turn to choose $\Delta$ and $w$. Note that both of them will affect the norm of $g$. To guarantee $\norm{g}_{{\mh}} \le B$, it has been shown~\cite{scarlett2017lower} that, when $\frac{\Delta}{B}$ is sufficient small, one can choose $w = \frac{2\zeta\pi l}{\sqrt{\log \frac{B(2\pi l^2)^{d/4}h(0)}{2\Delta} }}$ for $k_{\text{SE}}$ and $w = 2\zeta \left(2\Delta \frac{(8\pi^2)^{(\nu+d/2)/2}}{Bc^{-1/2}h(0)}\right)^{1/\nu}$ for $k_{\text{Mat\'ern}}$, where $c$ is an absolute constant. Note that, we will later consider $\Delta$ such that $\frac{\Delta}{B}$ is sufficient small.
	\item From the above choices of $w$ and $M = \lfloor\left(\frac{1}{w}\right)^d \rfloor$, we have  
	\begin{lm}
		\begin{align}
		\label{eq:M}
			M = \Theta\left( \left(\ln \frac{B}{\Delta}\right)^{d/2} \right), \text{ for } k_{\text{SE}}, \text{ and } M = \Theta\left(\left(\frac{B}{\Delta}\right)^{d/\nu}\right) \text{ for } k_{\text{Mat\'ern}}
		\end{align}
	\end{lm}
	Since we choose a sufficient small $\frac{\Delta}{B}$, $M \gg 1$, i.e., there is an enough number of functions in $\mathcal{F}$.
	\item Finally, we choose $f$ as a uniformly sampled function from $\mathcal{F}$.

\end{itemize}

\subsection{Construction of reward function}
We consider the general case that the non-private reward (i.e., before passed to Laplace mechanism) could be heavy-tailed. That is, the $(1+\alpha)$th moment of the reward is bounded by some value $v$ for $\alpha \in (0,1]$. To satisfy this condition, we consider the same reward function in~\cite{chowdhury2019bayesian}, i.e., 
\begin{lm}
\begin{align}
\label{eq:reward_dist}
	y(x) =
	    \begin{cases*}
	     sgn(f(x))\left(\frac{v}{2\Delta}\right)^{1/\alpha} & with probability $\left(\frac{2\Delta}{v}\right)^{\frac{1}{\alpha}}|f(x)|$ \\
	     0 & otherwise
	    \end{cases*}
\end{align}	
\end{lm}
Note that as long as $\Delta \le \frac{1}{2}v^{\frac{1}{1+\alpha}}$, Eq.~\eqref{eq:reward_dist} is a valid distribution.

\subsection{Preliminary notations and lemmas}
Now, let us introduce some necessary notations and results. In the following, with a bit abuse of notations, we use $\tilde{P}$ (probability) (or $\tilde{E}$, expectation) for the randomness with respect to non-private rewards, and $P$ (or $E$) for the private rewards. 
\begin{itemize}
	\item $y_m$ denotes the non-private reward (i.e., before the Laplace mechanism) when the underlying ground function is $f_m$. Let $\bar{y}_m$ denote the corresponding private reward. $f_0$ denotes the function which has value zero everywhere, and $y_0$ (resp. $\bar{y}_0$) is the corresponding non-private (private) reward. Let $\bar{Y}_T = \{\bar{y}_1,\ldots,\bar{y}_T\}$ be the sequence of private rewards up to time $T$. Let $P_m$ (resp. $P_0$ ) be the probability density function of the private reward sequence $\bar{Y}_T$ when the underlying function is $f_m$ (resp. $f_0$). Similarly, $\tilde{P}_m$ (resp. $\tilde{P}_0$) for the non-private sequence $Y_T$. Further, $P_m(\bar{y}\mid x)$ (resp. $P_0(\bar{y}\mid x)$) denotes the conditional probability of the private reward $\bar{y}$ given the selected point $x$ when the underlying function is $f_m$ (resp. $f_0$). Similarly, $\tilde{P}_m({y}\mid x)$ (resp. $\tilde{P}_0({y}\mid x)$) for the non-private reward $y$.
	\item $\mathbb{E}_m$ (resp. $\mathbb{E}_0$) denotes expectation (with respect to the noisy private rewards) when the underlying function is $f_m$ (resp. $f_0$). $\mathbb{E}[\cdot] = \frac{1}{M}\sum_{m=1}^M\mathbb{E}_m[\cdot]$ denotes the expectation averaged over a uniformly random function index.
	\item $\{\mathcal{R}_m\}_{m=1}^M$ denote a partition of the domain into $M$ regions (i.e., the uniform grid) with $f_m$ taking its maximum in the center of $\mathcal{R}_m$. Define the maximum absolute value of $f_m$ in the region $\mathcal{R}_j$ as
	\begin{lm}
		\begin{align}
		\label{eq:def_vmj}
			{v}_m^j :=\max_{x\in \mathcal{R}_j} |f_m(x)|.
		\end{align}
	\end{lm}
	Define the maximum KL divergence between $P_0(\cdot | x)$ and $P_m(\cdot | x)$ within region $\mathcal{R}_j$ as
	\begin{lm}
		\begin{align}
		\label{eq:def_dmj}
			D_m^j:=\max_{x\in \mathcal{R}_j}D_{kl}(P_0(\cdot|x) || P_m(\cdot|x) )
		\end{align}
	\end{lm}
	\item $N_j = \sum_{t=1}^T \mathbbm{1}\{x_t \in \mathcal{R}_j\}$ denotes the number of points within $\mathcal{R}_j$ that are selected up to time $T$.
	 \end{itemize}

	Next, we provide some useful results for our LDP scenario. The first result is directly from Lemma 5 of~\cite{scarlett2017lower}.

	\begin{lemma}
	\label{lem:vmj}
		The function class $\mathcal{F}$ constructed in~\ref{sec:construction} satisfies the following properties:
		\begin{enumerate}
			\item $\sum_{j=1}^M v_m^j = O(\Delta)$ for all $m \in [M]$.
			\item $\sum_{m=1}^M v_m^j = O(\Delta)$ for all $j \in [M]$.
		\end{enumerate}
	\end{lemma}

	\begin{lemma}
	\label{lem:EN}
		Under the definitions above, we have $\mathbb{E}_m(N_j) \le \mathbb{E}_0(N_j) + T\sqrt{\sum_{j=1}^M\mathbb{E}_0[N_j]D_m^j}$
	\end{lemma}
	\begin{proof}
		First, by Lemma 3 of~\cite{scarlett2017lower}, which relates two expectations in terms of the corresponding divergence of the probability distributions, we have 
		\begin{lm}
			\begin{align*}
				\mathbb{E}_m\left[f(\bar{Y}_T)\right] \le \mathbb{E}_0\left[f(\bar{Y}_T)\right] + A\sqrt{D_{kl}(P_0 || P_m) },
			\end{align*}
		\end{lm}
		for any function $f$ on $\bar{Y}_T$ with a bounded range $[0,A]$. Note that compared to the non-private case in~\cite{scarlett2017lower}, here the randomness is with respect to the private reward sequences. Then, following Lemma 4 in~\cite{scarlett2017lower}, we have 
		\begin{lm}
			\begin{align*}
				D_{kl}(P_0 || P_m) \le \sum_{j=1}^M\mathbb{E}_0[N_j]D_m^j.
			\end{align*}
		\end{lm}
		Further, note that $\{x_t\}_{t=1}^T$ is a function of $\bar{Y}_T$.  Combing the two results above, yields the result.
	\end{proof}

	\subsection{Analysis of expected cumulative regret }
	Note that $D_m^j$ is the key term in our analysis. In particular, in our LDP setting, it is the KL divergence between two distributions on the \emph{private} rewards, which is the key difference compared to previous works. Nevertheless, as stated before, we can relate it to the non-private case. Specifically, we have the following bound on it.
	\begin{lemma}
	\label{lem:dmj}
		Under the definitions above, if $\Delta \le \frac{1}{2}\left(\frac{1}{2}\right)^{\frac{\alpha}{1+\alpha} }v^{\frac{1}{1+\alpha} }$, we have, for any $\epsilon$-LDP mechanism
		\begin{lm}
			\begin{align*}
				D_m^j \le 8(e^{\epsilon}-1)^2 2^{\frac{1+\alpha}{\alpha}}\left(\frac{\Delta}{v}\right)^{\frac{1}{\alpha} } v_m^j.
			\end{align*}
		\end{lm}
	\end{lemma}
	\begin{proof}
	First, we have 
		\begin{lm}
			\begin{align*}
				D_{kl}(P_0(\cdot|x) || P_m(\cdot|x) ) + D_{kl}(P_m(\cdot|x) || P_0(\cdot|x) ) &\lep{a} 4(e^{\epsilon}-1)^2 ||\tilde{P}_0(\cdot|x) - \tilde{P}_m(\cdot|x) ||^2_{TV}\\
				&\lep{b} 8(e^{\epsilon}-1)^2 D_{kl}(\tilde{P}_0(\cdot|x) || \tilde{P}_m(\cdot|x) ),
			\end{align*}
		\end{lm}
		where (a) follows from Theorem 1 of~\cite{duchi2013local} and (b) is due to Pinsker's inequality. Then, according to the (non-private) reward distribution in Eq.~\eqref{eq:reward_dist} and the analysis in~\cite{chowdhury2019bayesian} (cf., Eq. (17)), we have 
		\begin{lm}
			\begin{align*}
				D_{kl}(\tilde{P}_0(\cdot|x) || \tilde{P}_m(\cdot|x) ) \le 2\left(\frac{2\Delta}{v}\right)^{\frac{1}{\alpha}}|f(x)|
			\end{align*}
		\end{lm}
		for $\Delta \le \frac{1}{2}\left(\frac{1}{2}\right)^{\frac{\alpha}{1+\alpha} }v^{\frac{1}{1+\alpha} }$. Thus, combing the results above with the definitions of $D_m^j$ and $v_m^j$ in Eqs.~\eqref{eq:def_dmj} and~\eqref{eq:def_vmj}, yields the result. 
	\end{proof}

	Now, we are well-prepared to analyze the lower bounds of the expected cumulative regret in the LDP setting.

	First, note that by the decomposition of expectation, we have $\mathbb{E}_m[f(x_t)] \le \sum_{j=1}^M \mathbb{P}_m[x_t \in \mathcal{R}_j] v_m^j$. Hence, 
	\begin{lm}
		\begin{align*}
			\exm{\sum_{t=1}^T f(x_t) } \le \sum_{j=1}^M v_m^j \exm{N_j} \lep{a} \sum_{j=1}^M v_m^j\left(\mathbb{E}_0(N_j) + T\sqrt{\sum_{j'=1}^M\mathbb{E}_0[N_{j'}]D_m^{j'} }\right),
		\end{align*}
	\end{lm}
	where (a) directly follows from Lemma~\ref{lem:EN}. Now, taking the average over $m \in [M]$, yields
	\begin{lm}
		\begin{align}
		\label{eq:total_lb}
			\ex{\sum_{j=1}^T f(x_t)} \le \frac{1}{M}\sum_{m=1}^M\sum_{j=1}^M v_m^j\left(\mathbb{E}_0(N_j) + T\sqrt{\sum_{j'=1}^M\mathbb{E}_0[N_{j'}]D_m^{j'} }\right)
		\end{align}
	\end{lm}
	For the first term, we can bound it as follows:
	\begin{lm}
		\begin{align}
		\label{eq:first_lb}
			\frac{1}{M}\sum_{m=1}^M\sum_{j=1}^M v_m^j \mathbb{E}_0[N_j] = \frac{1}{M}\sum_{j=1}^M\sum_{m=1}^M v_m^j \mathbb{E}_0[N_j] \lep{a} O\left(\frac{\Delta}{M}\right) \sum_{j=1}^M\exo{N_j} \ep{b} O\left(\frac{T\Delta}{M}\right),
		\end{align}
	\end{lm}
	where (a) follows from the second part of Lemma~\ref{lem:vmj}, and (b) holds since $\sum_{j=1}^M N_j = T$.

	For the second term, we have when $\Delta \le \frac{1}{2}\left(\frac{1}{2}\right)^{\frac{\alpha}{1+\alpha} }v^{\frac{1}{1+\alpha} }$
	\begin{lm}
		\begin{align}
			\frac{1}{M}\sum_{m=1}^M\sum_{j=1}^M v_m^j T\sqrt{\sum_{j'=1}^M\mathbb{E}_0[N_{j'}]D_m^{j'} } &\ep{a} O(T\Delta) \frac{1}{M} \sum_{m=1}^M \sqrt{\sum_{j'=1}^M\mathbb{E}_0[N_{j'}]D_m^{j'} }\nonumber\\
			&\lep{b}O(T\Delta) \sqrt{\frac{1}{M} \sum_{m=1}^M\sum_{j'=1}^M\mathbb{E}_0[N_{j'}]D_m^{j'} }\nonumber\\
			&\lep{c}O(T\Delta)(e^{\epsilon}-1)2^{\frac{1+\alpha}{2\alpha}}\left(\frac{\Delta}{v}\right)^{\frac{1}{2\alpha} }\sqrt{\frac{1}{M} \sum_{m=1}^M\sum_{j'=1}^M\mathbb{E}_0[N_{j'}]v_m^{j'} }\nonumber\\
			&\lep{d} O\left(T \Delta (e^{\epsilon}-1)\frac{(2\Delta)^{\frac{1+\alpha}{2\alpha} }}{v^{\frac{1}{2\alpha} }} \sqrt{\frac{T}{M}}\right),\label{eq:sec_lb}
		\end{align}
	\end{lm}
	where (a) follows from the first part of Lemma~\ref{lem:vmj}; (b) follows from Jensen's inequality; (c) follows from Lemma~\ref{lem:dmj}; (d) holds by Eq.~\eqref{eq:first_lb}. Now, substituting Eqs.~\eqref{eq:first_lb} and~\eqref{eq:sec_lb} into Eq.~\eqref{eq:total_lb}, yields
	\begin{lm}
		\begin{align*}
			\ex{\sum_{j=1}^T f(x_t)} \le  CT\Delta\left(\frac{1}{M} + (e^{\epsilon}-1)\frac{(2\Delta)^{\frac{1+\alpha}{2\alpha} }}{v^{\frac{1}{2\alpha} }} \sqrt{\frac{T}{M}}\right)
		\end{align*}
	\end{lm}
	for some constant $C$, when $\Delta \le \frac{1}{2}\left(\frac{1}{2}\right)^{\frac{\alpha}{1+\alpha} }v^{\frac{1}{1+\alpha} }$.

	Observe that $f(x^*) = 2\Delta$ by the construction process in~\ref{sec:construction}, then the cumulative expected regret is given by
	\begin{lm}
		\begin{align*}
			\ex{R_T} = Tf(x^*) - \ex{\sum_{j=1}^T f(x_t)} \ge T\Delta\left(2 - \frac{C}{M} - C(e^{\epsilon}-1)\frac{(2\Delta)^{\frac{1+\alpha}{2\alpha} }}{v^{\frac{1}{2\alpha} }} \sqrt{\frac{T}{M}}\right),
		\end{align*}
	\end{lm}
	when $\Delta \le \frac{1}{2}\left(\frac{1}{2}\right)^{\frac{\alpha}{1+\alpha} }v^{\frac{1}{1+\alpha} }$. Recall that we always choose a sufficient $\frac{\Delta}{B}$, and hence by Eq.~\eqref{eq:M}, $M$ is sufficient large. Thus, $\frac{C}{M} \le \frac{1}{2}$. As a result, we have 
	\begin{lm}
		\begin{align*}
			\ex{R_T} &\ge T\Delta\left(\frac{3}{2} - C(e^{\epsilon}-1)\frac{(2\Delta)^{\frac{1+\alpha}{2\alpha} }}{v^{\frac{1}{2\alpha} }} \sqrt{\frac{T}{M}}\right)\\
			&\ge T\Delta, \text{ for } \Delta \le \frac{1}{2}\left(\min\left\{\frac{1}{2}, \frac{M}{4C^2(e^{\epsilon}-1)^2T}\right\}\right)^{\frac{\alpha}{1+\alpha} }v^{\frac{1}{1+\alpha}}.
		\end{align*}
	\end{lm}
	Therefore, if $M\le 2C^2(e^{\epsilon}-1)^2T$, then we have 
	\begin{lm}
		\begin{align}
		\label{eq:RT_lb}
			\ex{R_T} = \Omega\left(v^{\frac{1}{1+\alpha}} T^{\frac{1}{1+\alpha}} M^{\frac{\alpha}{1+\alpha}}  (e^{\epsilon}-1)^{-\frac{2\alpha}{1+\alpha} } \right) \text{ for } \frac{1}{4}\left(\frac{M}{4C^2(e^{\epsilon}-1)^2T}\right)^{\frac{\alpha}{1+\alpha} }v^{\frac{1}{1+\alpha}} \le \Delta \le \frac{1}{2}\left(\frac{M}{4C^2(e^{\epsilon}-1)^2T}\right)^{\frac{\alpha}{1+\alpha} }v^{\frac{1}{1+\alpha}}
		\end{align}
	\end{lm}

	Now, it only remains to combine $M$ (in Eq.~\eqref{eq:M}) and Eq.~\eqref{eq:RT_lb} to arrive at the specific lower bound for $k_{\text{SE}}$ and $k_{\text{Mat\'ern}}$, respectively.

	\begin{itemize}
		\item For $k_{\text{SE}}$, note that we choose $M = \Theta\left( \left(\ln \frac{B}{\Delta}\right)^{d/2} \right)$ in Eq.~\eqref{eq:M}. Combing it with the upper and lower bounds on $\Delta$ in Eq.~\eqref{eq:RT_lb}, yields that $\Delta = \Theta\left( \left(\frac{\left(\ln \frac{B}{\Delta}\right)^{d/2}}{(e^{\epsilon}-1)^2T}\right)^{\frac{\alpha}{1+\alpha} }v^{\frac{1}{1+\alpha}}\right) $. This, in turn, implies that 
		\begin{lm}
			\begin{align*}
				\ln\frac{B}{\Delta} = \ln \frac{B\left(T(e^{\epsilon}-1)^2\right)^{\frac{\alpha}{1+\alpha} }}{v^{\frac{1}{1+\alpha}}} - \ln \left(\Theta(1)\left(\ln \frac{B}{\Delta}\right)^{\frac{d\alpha}{2(1+\alpha)} } \right).
			\end{align*}
		\end{lm}
		Since $d = O(1)$, and $\frac{\alpha}{1+\alpha} \in (0,1/2]$, the second term above behaves like $\Theta(\ln \ln \frac{B}{\Delta})$, which is at most $\frac{1}{2}\ln \frac{B}{\Delta}$ for sufficiently small $\frac{\Delta}{B}$. Thus, we have  $\ln\frac{B}{\Delta} = \Theta\left(\ln \frac{B\left(T(e^{\epsilon}-1)^2\right)^{\frac{\alpha}{1+\alpha} }}{v^{\frac{1}{1+\alpha}}}\right)$, which implies that 
		\begin{lm}
			\begin{align*}
				M = \Theta\left(\left(\ln \frac{B\left(T(e^{\epsilon}-1)^2\right)^{\frac{\alpha}{1+\alpha} }}{v^{\frac{1}{1+\alpha}}}\right)^{\frac{d}{2}}\right)
			\end{align*}
		\end{lm}
		and 
		\begin{lm}
			\begin{align*}
				\Delta = \Theta\left(\left(\ln \frac{B\left(T(e^{\epsilon}-1)^2\right)^{\frac{\alpha}{1+\alpha} }}{v^{\frac{1}{1+\alpha}}}\right)^{\frac{d\alpha}{2(1+\alpha)}} T^{-\frac{\alpha}{1+\alpha}} (e^{\epsilon}-1)^{-\frac{2\alpha}{1+\alpha}} v^{\frac{1}{1+\alpha}}\right).
			\end{align*}
		\end{lm}

	Note that the choice $M$ ensures that $M \le 2C^2(e^{\epsilon}-1)^2T$ and the choice of $\Delta$ guarantees that $\frac{\Delta}{B}$ is indeed sufficiently small as long as $v^{\frac{1}{1+\alpha}} \le C_1 B(T (e^{\epsilon}-1)^2)^{\frac{\alpha}{1+\alpha}}$ for some sufficiently small constant $C_1$. This can always hold since $B, v, \epsilon$ are all constants that do not increase with $T$. Finally, substituting $M$ into Eq.~\eqref{eq:RT_lb}, yields 
	\begin{lm}
		\begin{align*}
			\ex{R_T} &= \Omega\left(\left(\ln \frac{B\left(T(e^{\epsilon}-1)^2\right)^{\frac{\alpha}{1+\alpha} }}{v^{\frac{1}{1+\alpha}}}\right)^{\frac{d\alpha}{2(1+\alpha)}} T^{\frac{1}{1+\alpha}} (e^{\epsilon}-1)^{-\frac{2\alpha}{1+\alpha}} v^{\frac{1}{1+\alpha}}\right)\\
			&\ep{a}\Omega\left(\left(\ln \frac{B^{\frac{1+\alpha}{\alpha}}  T(e^{\epsilon}-1)^2}{v^{\frac{1}{\alpha}}}\right)^{\frac{d\alpha}{2(1+\alpha)}} T^{\frac{1}{1+\alpha}} (e^{\epsilon}-1)^{-\frac{2\alpha}{1+\alpha}} v^{\frac{1}{1+\alpha}}\right)
		\end{align*}
	\end{lm}
	where (a) holds due to $d = O(1)$ and $\frac{\alpha}{1+\alpha} \in (0,\frac{1}{2}]$.

	\item For $k_{\text{Mat\'ern}}$, from Eq.~\eqref{eq:M}, we have $M = \Theta\left(\left(\frac{B}{\Delta}\right)^{d/\nu}\right)$. Plug it into the upper and lower bounds on $\Delta$ in Eq.~\eqref{eq:RT_lb}, we have $\Delta = \Theta\left(\left(\frac{1}{T} \frac{1}{(e^{\epsilon}-1)^2} \left(\frac{B}{\Delta}\right)^{\frac{d}{\nu}}  \right)^{\frac{\alpha}{1+\alpha}} v^{\frac{1}{1+\alpha}}\right)$. This further implies that 
	\begin{lm}
		\begin{align*}
			\Delta = \Theta\left( v^{\frac{\nu/(1+\alpha)}{\nu+d\alpha/(1+\alpha)}} B^{\frac{d\alpha/(1+\alpha)}{\nu+d\alpha/(1+\alpha)}} (T (e^{\epsilon}-1)^2)^{-\frac{\nu\alpha/(1+\alpha)}{\nu+d\alpha/(1+\alpha)} } \right)
		\end{align*}
	\end{lm}
	and 
	\begin{lm}
		\begin{align*}
			M = \Theta\left(v^{-\frac{d/(1+\alpha)}{\nu+d\alpha/(1+\alpha)}} B^{\frac{d}{\nu+d\alpha/(1+\alpha)} }  (T (e^{\epsilon}-1)^2)^{\frac{d\alpha/(1+\alpha)}{\nu+d\alpha/(1+\alpha)} }  \right)
		\end{align*}
	\end{lm}
	As before, the choice of $M$ and $\Delta$ above can guarantee our necessary conditions. Finally, substituting $M$ into Eq.~\eqref{eq:RT_lb}, yields 
	\begin{lm}
		\begin{align*}
			\ex{R_T} = \Omega\left( v^{\frac{\nu}{\nu(1+\alpha)+d\alpha}}T^{\frac{\nu+d\alpha}{\nu(1+\alpha)+d\alpha}} {(e^{\epsilon}-1)^{\frac{-2\alpha}{1+\alpha}}} (e^{\epsilon}-1)^{\frac{2d\alpha^2}{(1+\alpha)(\nu(1+\alpha)+d\alpha)} } B^{\frac{d\alpha}{\nu(1+\alpha)+d\alpha}}  \right).
		\end{align*}
	\end{lm}
	\end{itemize}
	\qed

\section{Proof of Lemma~\ref{lem:CTL}}
\begin{proof}
	By the reproducing property of RKHS, we have $f(x) = \inner{f}{k(x,\cdot)}_{\mh} \le \norm{f}_{\mh}\sqrt{k(x,x)}\le B$ for any $x\in \mathcal{D}$. Combining this with the upper bound $R$ on the noise $\eta_t$, yields that the sensitivity of information is at most $2(B+R)$. Thus, $\epsilon$-LDP follows directly from the Laplace mechanism~\cite{dwork2014algorithmic}.
\end{proof}

\section{Proof of Theorem~\ref{thm:ATA}}
\begin{proof}
	Note that in our LDP setting with additional independent Laplace noise, $\ex{|\bar{y}_t|^{2} \mid \mathcal{F}_{t-1}}\le B^2+R^2+2\mathcal{L}^2$, where $\mathcal{L} = \frac{2(B+R)}{\epsilon}$. Hence, Theorem 4 of~\cite{chowdhury2019bayesian} is satisfied with $\alpha = 1$ and $v = B^2+R^2 +8(B+R)^2/\epsilon^2$, which directly leads to our results.
\end{proof}

\section{Proofs of Theorems~\ref{thm:TGP} and~\ref{thm:MoMA}}
\subsection{Preliminaries}
In this section, we first present some common notations and useful results for the proofs of the two theorems. 

Similar to~\cite{chowdhury2019bayesian}, we define $\varphi: \mathcal{D} \to \mathcal{H}_k(\mathcal{D})$ as the associated feature map of the kernel $k$ such that $k(x,y) = \inner{\varphi(x)}{\varphi(y)}_{\mh}$ for any $x, y \in \mathcal{D}$ From the reproducing property, we have for any $f \in \mathcal{H}_k(\mathcal{D})$, $f(x) = \inner{f}{\varphi(x)}_{\mh}$. For a set $\{x_1, \ldots, x_t\}$, we define a operator $\Phi_t: \mathcal{H}_k(\mathcal{D}) \to \mathbb{R}^t$ such that for any $f \in \mathcal{H}_k(\mathcal{D})$, $\Phi_t f = [\inner{\varphi(x_1)}{f}_{\mh},\ldots, \inner{\varphi(x_t)}{f}_{\mh}]^T$, which is also equal to $[f(x_1),\ldots, f(x_t)]^T$ by the reproducing property. Define the adjoint of $\Phi_t$ by $\Phi_t^T: \mathbb{R}^t \to \mathcal{H}_k(\mathcal{D})$. For any $\lambda > 0$, define $V_t = \Phi_t^T\Phi_t + \lambda {I}_{\mh}$, where $I_{\mh}$ denotes the identity operator. For a positive definite operator $V : \mathcal{H}_k(\mathcal{D}) \to \mathcal{H}_k(\mathcal{D})$, we define the inner product $\inner{\cdot}{\cdot}_V : =\inner{\cdot}{V\cdot}_{\mh}$ with the corresponding norm as $\norm{\cdot}_V$.

Based on the definitions above, we have the following result, which is directly from Eq.(13) of~\cite{chowdhury2017kernelized}.

\begin{lemma}
\label{lem:sigma_t}
	Under the definitions above, we have $\sigma_t^2(x) = \lambda \norm{\varphi(x)}_{V_t^{-1}}^2$.
\end{lemma}

The following lemma summarizes useful properties for linear operators, which is from Lemma 3 of~\cite{chowdhury2019bayesian}.
\begin{lemma}
	For any linear operator $A: \mathcal{H}_k(\mathcal{D}) \to \mathbb{R}^t$ and its adjoint $A^T: \mathbb{R}^t \to \mathcal{H}_k(\mathcal{D})$, for any $\lambda > 0$, we have
	\begin{lm}
		\begin{align}
		\label{eq:op1}
			(A^TA + \lambda I_{\mh})^{-1} A^T = A^T(AA^T + \lambda I_t)^{-1}
		\end{align}
	\end{lm}
	and 
	\begin{lm}
		\begin{align}
		\label{eq:op2}
			I_{\mh} - A^T(AA^T+\lambda I_t)A = \lambda(A^TA+\lambda I_{\mh})^{-1}.
		\end{align}
	\end{lm}
\end{lemma}

The following result is a standard result in BO, which can be obtained from Lemma 6 of~\cite{chowdhury2019bayesian} and Cauchy-Schwartz inequality.
\begin{lemma}
\label{lem:sum_sigma}
	If $k(x,x) \le 1$, for any $x \in \mathcal{D}$, we have
	\begin{lm}
		\begin{align*}
			\sum_{t=1}^T \sigma_{t-1}(x) = O(\sqrt{\gamma_T T}).
		\end{align*}
	\end{lm}
\end{lemma}

\subsection{Proof of Theorem~\ref{thm:TGP}}
\begin{proof}
	First, as pointed out in the proof sketch, we establish a bound on $|f(x) - \hat{\mu}_t(x)|$, in which $\hat{\mu}_t(x) = k_t(x)^T(K_t + \lambda I)^{-1}\hat{Y}_t$. Following~\cite{chowdhury2019bayesian}, we define $\hat{\eta}_t = \hat{y}_t - f(x_t)$, $\hat{N}_t = [\hat{\eta}_1, \ldots, \hat{\eta}_t]^T$ and $f_t = [f(x_1),\ldots, f(x_t)]^T$. Hence, $\hat{Y}_t = \hat{N}_t + f_t$. Define $\alpha_t(x)=k_t(x)^T(K_t + \lambda I)^{-1}f_t$, then by the preliminaries above, we have 
	\begin{lm}
		\begin{align*}
			f(x) - \alpha_t(x) = \inner{\varphi(x)}{(I_{\mh}- \Phi_t^T(\Phi_t\Phi_t^T + \lambda I_t)^{-1}\Phi_t)f }_{\mh} \ep{a} \lambda\inner{\varphi(x)}{f}_{V_t^{-1}}
		\end{align*}
	\end{lm}
	where (a) holds by Eq.~\eqref{eq:op2}. Similarly, we have 
	\begin{lm}
		\begin{align*}
			k_t(x)^T(K_t + \lambda I)^{-1}\hat{N}_t = \inner{\varphi(x)}{\Phi_t^T(\Phi_t\Phi_t^T + \lambda I_t)^{-1}\hat{N}_t}_{\mh} = \inner{\varphi(x)}{\Phi_t^T \hat{N}_t}_{V_t^{-1}},
		\end{align*}
	\end{lm}
	which holds by Eq.~\eqref{eq:op1}.
	Based on the results above, we have 
	\begin{lm}
		\begin{align*}
			|f(x) - \hat{\mu}_t(x)| &\le |\lambda\inner{\varphi(x)}{f}_{V_t^{-1}}| + |\inner{\varphi(x)}{\Phi_t^T \hat{N}_t}_{V_t^{-1}}|\\
			&\lep{a} \lambda \norm{\varphi(x)}_{V_t^{-1}} \norm{V_t^{-1/2}f}_{\mh} + \norm{\varphi(x)}_{V_t^{-1}}\norm{\Phi_t^T\hat{N}_t}_{V_t^{-1}}\\
			&\lep{b} \lambda^{1/2}\norm{\varphi(x)}_{V_t^{-1}}\norm{f}_{\mh} + \norm{\varphi(x)}_{V_t^{-1}}\norm{\Phi_t^T\hat{N}_t}_{V_t^{-1}}\\
			&\lep{c} \sigma_t(x) \left(B +  \lambda^{-1/2}\norm{\Phi_t^T\hat{N}_t}_{V_t^{-1}}\right)
		\end{align*}
	\end{lm}
	where (a) follows from Cauchy-Schwartz inequality; (b) is true since $V_t^{-1} \preceq \lambda^{-1}I_{\mh}$; (c) follows from the fact that $\norm{f}_{\mh} \le B$ and Lemma~\ref{lem:sigma_t}.

	The key term is $ \norm{\Phi_t^T\hat{N}_t}_{V_t^{-1}} = \norms{\sum_{\tau = 1}^t\hat{\eta}_{\tau}\varphi(x_{\tau})}_{V_t^{-1}}$, which can be handled by the self-normalized inequality if $\hat{\eta}_{\tau}$ is sub-Gaussian. However, in our LDP setting, it is not due to Laplace noise. To overcome this issue, we will divide it into two parts. In particular, similar to~\cite{chowdhury2019bayesian}, we define $\xi_t = \hat{\eta}_t - \ex{\hat{\eta}_t \mid \mathcal{F}_{t-1}}$. Now, the key term can be written as
\begin{linenomath}
	\begin{align*}
 	\norms{\sum_{\tau = 1}^t\hat{\eta}_{\tau}\varphi(x_{\tau})}_{V_t^{-1}} = \underbrace{\norms{\sum_{\tau = 1}^t {\xi}_{\tau}\phi(x_{\tau})}_{V_t^{-1}} }_{\mathcal{T}_1} + \underbrace{\norms{\sum_{\tau = 1}^t \ex{\hat{\eta}_{\tau} | \mathcal{F}_{\tau-1}}\phi(x_{\tau})}_{V_t^{-1}}.}_{\mathcal{T}_2}
 \end{align*} 
\end{linenomath}
 For $\mathcal{T}_1$, note that $\xi_t = \hat{y}_t -\ex{\hat{y}_t \mid \mathcal{F}_{t-1}}$ and $\hat{y}_t = \bar{y}_t\mathbbm{1}_{|\bar{y}_t| \le b_t}$. This implies that $\xi_t$ is bounded by $2b_t$, and hence sub-Gaussian. Thus, by the self-normalized inequality for the RKHS-valued process in~\cite{durand2018streaming,chowdhury2019bayesian}, we have for any $\delta \in (0,1]$, with probability at least $1-\delta$, we have for all $t \in \mathbb{N}$
 \begin{linenomath}
 	\begin{align*}
 	\mathcal{T}_1 \le 2b_t\sqrt{2(\gamma_t + \ln(1/\delta))}.
 \end{align*}
 \end{linenomath}
 For $\mathcal{T}_2$, we can first bound it as  $\mathcal{T}_2 \le \sqrt{\sum_{\tau=1}^t \ex{\hat{\eta}_{\tau} | \mathcal{F}_{\tau-1}}^2}$. This is due to the following property. For any $a \in \mathbb{R}^t$, 
 \begin{lm}
 	\begin{align*}
 		\norm{\sum_{\tau = 1}^t a_{\tau} \varphi(x_{\tau})}_{V_t^{-1}}^2 = a^T\Phi_t(\Phi_t^T\Phi_t + \lambda I_{\mh})^{-1}\Phi_t^Ta \ep{a} a^T\Phi_t\Phi_t^T(\Phi_t\Phi_t^T+\lambda I_t)^{-1}a \lep{b} \norm{a}_2^2,
 	\end{align*}
 \end{lm}
 where (a) holds by Eq.~\eqref{eq:op1}; (b) is true since $\Phi_t\Phi_t^T(\Phi_t\Phi_t^T+\lambda I_t)^{-1}  \preceq  I_t$.

 Further, note that $\ex{\hat{\eta}_{\tau} | \mathcal{F}_{\tau-1}} = -\ex{\bar{y}_{\tau}\mathbbm{1}_{|\tilde{y}_{\tau}| > b_{\tau}}|\mathcal{F}_{\tau-1}}$. Hence, by Cauchy-Schwartz inequality with $b_{\tau} = B+R+\mathcal{L}\ln \tau$, we have 
 \begin{linenomath}
 	\begin{align*}
 	\ex{\hat{\eta}_{\tau} | \mathcal{F}_{\tau-1}}^2 \le \ex{\bar{y}_{\tau}^2|\mathcal{F}_{\tau-1}} \mathbbm{P}(|L| > \mathcal{L}\ln \tau) \le K\frac{1}{\tau},
 \end{align*}
 \end{linenomath}
where $K:=B^2+R^2+2\mathcal{L}^2$. The last inequality holds since $|L| \sim \text{Exp}(1/\mathcal{L})$. Therefore, by the property of Harmonic sum, we have 
\begin{linenomath}
	\begin{align*}
	\mathcal{T}_2 \le\sqrt{K(\ln t + 1)}.
\end{align*}
\end{linenomath}
Hence, by setting
\begin{linenomath}
	\begin{align*}
	\beta_{t+1} = B + \frac{2\sqrt{2}}{\sqrt{\lambda}}b_{t}\sqrt{\gamma_{t} + \ln(1/\delta)}+\frac{1}{\sqrt{\lambda}}\sqrt{K(\ln t+1)},
\end{align*}
\end{linenomath}
we have for any $\delta \in (0,1]$, with probability at least $1-\delta$, uniformly over all $t\ge 1$ and $x \in \mathcal{D}$,
\begin{lm}
	\begin{align}
	\label{eq:high-prob-TGP}
	|f(x) - \hat{\mu}_t(x)| \le \beta_{t+1}\sigma_t(x).
	\end{align}
\end{lm}
Now, the following analysis is standard. First,
\begin{lm}
	\begin{align*}
		r_t = f(x^*) - f(x_t) \le 2\beta_t\sigma_{t-1}(x),
	\end{align*}
\end{lm}
which follows from Eq.~\eqref{eq:high-prob-TGP} and the GP-UCB like algorithm. Hence, for any $\delta \in (0,1]$, with probability at least $1-\delta$, the cumulative regret is given by 
\begin{lm}
	\begin{align*}
		R_T = \sum_{t=1}^T r_t \le 2\beta_T \sum_{t=1}^T\sigma_{t-1}(x) \ep{a} O(\beta_T \sqrt{\gamma_T T}) = O\left(\vartheta \sqrt{\ln T\gamma_T T} + \vartheta\sqrt{T}\ln T\sqrt{\gamma_T (\gamma_T + \ln(1/\delta))}\right),
	\end{align*}
\end{lm}
where $\vartheta = (B+R)/\epsilon$ and (a) holds by Lemma~\ref{lem:sum_sigma}.
\end{proof}

\subsection{Proof of Theorem~\ref{thm:MoMA}}
Before the proof, we first present some useful results regarding Nystr\"{o}m approximation based on~\cite{calandriello2019gaussian,chowdhury2019bayesian}. One key notation difference here is that in MoMA-GP-UCB, the main index  is $n$ (i.e., epoch) rather than $t$, since the Nystr\"{o}m approximation is only updated every epoch, which consists of $k$ iterations.

First, we present the algorithm for Nystr\"{o}mEmbedding used in MoMA-GP-UCB based on~\cite{chowdhury2019bayesian}, as shown in Algorithm~\ref{alg:embed}. The basic idea is to first sample a random number $m_n$ of points from the set $\{x_1,\ldots,x_n\}$ to construct a dictionary $\mathcal{D}_n = \{x_{i_1},\ldots, x_{i_{m_n}} \}$, $i_j \in [n]$. Each data point $x_i$, $i\in[n]$ is included with probability $p_{n,i}=\min\{q \tilde{\sigma}_{n-1}^2(x_i)),1 \}$. Then, the finite approximation of the feature map is computed in line $7$. Note that $A^{\dagger}$ denotes the pseudo inverse of any matrix $A$. $K_{\mathcal{D}_n} = [k(u,v)]_{u,v \in \mathcal{D}_n}$ and $k_{\mathcal{D}_n}(x) = [k(x_{i_1}, x),\ldots, k(x_{i_{m_n}}, x)]^T$.
\begin{algorithm}[t]
\caption{Nystr\"{o}mEmbedding}\label{alg:embed}
\begin{algorithmic}[1]
\State \textbf{Input:} $\{(x_i,\tilde{\sigma}_{t-1}(x_i))\}_{i=1}^t$, $q$
\State \textbf{Set:} $\mathcal{D}_0 = \emptyset$
        \For{{$t = 1,2,3,\ldots, t$}}
        	\State Sample $z_{t,i} \sim \text{Bern}(\min\{q \tilde{\sigma}_{t-1}^2(x_i)),1 \})$
        	\State If $z_{t,i} = 1$, then set $\mathcal{D}_t = \mathcal{D}_t \cup \{x_i\}$
		\EndFor
\State \textbf{Return} $\tilde{\varphi}_t(x) = \left(K_{\mathcal{D}_t}^{1/2} \right)^{\dagger} k_{\mathcal{D}_t}(x)$
\end{algorithmic}
\end{algorithm}

The following result states one equivalent definition of an $\epsilon$-accurate dictionary $\mathcal{D}_n$, based on Lemma 16 of~\cite{chowdhury2019bayesian}.
\begin{definition}[$\varepsilon$-accurate]
	For any $\varepsilon \in (0,1)$, a dictionary $\mathcal{D}_n \subseteq \{x_1,\ldots,x_n\}$ is said to be $\varepsilon$-accurate if $V_{\mathcal{D}_n}:=\Phi_n^TS_n^2\Phi_n + \lambda I_{\mh}$ satisfies 
	\begin{lm}
		\begin{align*}
			(1-\epsilon)V_n \preceq V_{\mathcal{D}_n} \preceq (1+\epsilon)V_n
		\end{align*}
	\end{lm}
	where $\mathcal{S}_n$ is the selection matrix associated with the dictionary $\mathcal{D}_n$ such that $[S_n]_{i,i} = 1/\sqrt{p_{n,i}}$ if $x_i \in \mathcal{D}_n$ and $0$, elsewhere.
\end{definition}

The following results states the key properties of Nystr\"{o}mEmbedding, which is directly adapted from Lemma 17 of~\cite{chowdhury2019bayesian}.
\begin{lemma}
\label{lem:events}
	For any $\varepsilon \in (0,1)$ and $\delta \in (0,1]$, set $\rho = \frac{1+\varepsilon}{1-\varepsilon}$ and $q = \frac{6\rho\ln(2T/\delta)}{\varepsilon^2}$. Let $E_{1,n}$ denote the event that $\mathcal{D}_n$ is $\varepsilon$-accurate, and $E_{2,n}$ denote the event that $m_n \le 6\rho\left(1+\frac{1}{\lambda}\right)q\gamma_t$. Then, with probability at least $1-\delta$, uniformly over all $ n\in [N]$, both $E_{1,n}$ and $E_{2,n}$ are true, i.e., $\mathbb{P}[\cap_{n=1}^N (E_{1,n}\cup E_{2,n} )] \ge 1-\delta$.
\end{lemma}
We also let $\mathcal{G}_n = \sigma\left( \{x_i, (z_{i,j})_{j=1}^i\}_{i=1}^n  \right)$, $n\ge1$ denote the $\sigma$-algebra generated by the arms played and the outcomes of the Nystr\"{o}mEmbedding algorithm up to time $t$. Note that $(\mathcal{G}_n)_{n\ge1}$ defines a filtration and both $E_{1,n}$ and $E_{2,n}$ are $(\mathcal{G}_n)_{n\ge1}$ measurable.

Based on the definitions above, we have the following results, which are useful for establishing the high probability confidence bound. The first result is adapted from the proof of Lemma 20 in~\cite{chowdhury2019bayesian} (cf. Eq.(25) and (26)), and the second one directly follows from Lemma 19 of~\cite{chowdhury2019bayesian}.
\begin{lemma}
\label{lem:CI_MoMA_pre}
	Given a filtration $(\mathcal{G}_n)_{n\ge1}$ such that $E_{1,n}$ is true (i.e., $\varepsilon$-accurate), we have
	\begin{enumerate}
		\item $|f(x) - \tilde{\mu}_n(x)| \le B\left(1+\frac{1}{\sqrt{1-\varepsilon}}\right)\tilde{\sigma}_{n}(x) + \lambda^{-1/2}\norms{\tilde{V}_n^{-1}\tilde{\Phi}_n^Tf_n - \tilde{\theta}_{n,k^*} }_{\tilde{V}_n} \tilde{\sigma}_{n}(x).$
		\item $\tilde{\sigma}_{n}(x) \le \rho \sigma_n(x)$, where $\rho = \frac{1+\varepsilon}{1-\varepsilon}$.
	\end{enumerate}	
\end{lemma}

The following two lemmas state the key properties of the least-square-estimators (LSEs) constructed in MoMA-GP-UCB, which are mainly inspired from~\cite{shao2018almost}.

\begin{lemma}
\label{lem:LSE_j}
	Let $\gamma := (9m_n c)^{\frac{1}{1+\alpha}}n^{\frac{1-\alpha}{2(1+\alpha)} }$, for any $j \in [k]$, we have
	\begin{linenomath}
		\begin{align*}
		\mathbb{P}\left( \norm{\tilde{V}_n^{-1}\tilde{\Phi}_n^Tf_n - \tilde{\theta}_{n,j} }_{\tilde{V}_n} \le  \gamma\right) \ge \frac{3}{4}.
	\end{align*}
	\end{linenomath}
\end{lemma}
\begin{proof}
	Recall that in MoMA-GP-UCB, $\tilde{\theta}_{n,j} = \tilde{V}_n^{-1}\sum_{i=1}^n y_{i,j}\tilde{\varphi}_n(x_i) = \tilde{V}_n^{-1}\tilde{\Phi}_n^TY_n$, where $Y_n = [y_{1,j}, y_{2,j}, \ldots, y_{n,j}]^T$.	Hence, we have 
	\begin{lm}
		\begin{align*}
			\norm{\tilde{V}_n^{-1}\tilde{\Phi}_n^Tf_n - \tilde{\theta}_{n,j} }_{\tilde{V}_n} = \norm{\tilde{V}_n^{-1}\tilde{\Phi}_n^Tf_n - \tilde{V}_n^{-1}\tilde{\Phi}_n^TY_n }_{\tilde{V}_n} = \norm{\tilde{V}_n^{-\frac{1}{2}} \tilde{\Phi}_n^T (Y_n -f_n)}_2 = \sqrt{\sum_{i=1}^{m_n}(u_i^T(Y_n-f_n))^2 },
		\end{align*}
	\end{lm}
	where $u_i^T$, $i\in[m_n]$ are the rows of $\tilde{V}_n^{-\frac{1}{2}} \tilde{\Phi}_n^T$. By Lemma 12 in~\cite{chowdhury2019bayesian}, we can bound the $(1+\alpha)$th norm. Specifically, we have $\norm{u_i}_{1+\alpha} \le n^{\frac{1-\alpha}{2(1+\alpha)} }$. In the following, we let $\eta_{l,j} = y_{l,j} - f(x_l)$.

	Then, we will follow a similar analysis as in~\cite{shao2018almost} (cf. Lemma 2). First, we bound the following probability
	\begin{lm}
		\begin{align*}
			\mathbb{P}\left(\sum_{i=1}^{m_n}\left(\sum_{\tau=1}^n u_{i,\tau}\eta_{\tau,j} \right)^2  >\gamma^2 \right) \le \underbrace{\mathbb{P}\left(\exists i, \tau, |u_{i,\tau}\eta_{\tau,j}| > \gamma \right)}_{\mathcal{T}_1} + \underbrace{\mathbb{P}\left(\sum_{i=1}^{m_n} \left(\sum_{\tau=1}^n u_{i,\tau}\eta_{\tau,j}\mathbbm{1}_{|u_{i,\tau}\eta_{\tau,j}| \le \gamma}  \right)^2 > \gamma^2\right)}_{\mathcal{T}_2}.
		\end{align*}
	\end{lm}
	For $\mathcal{T}_1$, by using union bound and Markov's inequality along with the norm bound on $u_i$, we have 
	\begin{lm}
		\begin{align*}
			\mathcal{T}_1 \le \frac{\sum_{i=1}^{m_n}\sum_{\tau=1}^n \ex{|u_{i,\tau}\eta_{\tau,j} |^{1+\alpha}}  }{\gamma^{1+\alpha}} \le \frac{\sum_{i=1}^{m_n}\sum_{\tau=1}^n |u_{i,\tau}|^{1+\alpha} c}{\gamma^{1+\alpha}} \le \frac{m_n c n^{\frac{1-\alpha}{2} }}{\gamma^{1+\alpha}}.
		\end{align*}
	\end{lm}
	For $\mathcal{T}_2$, by using Markov's inequality and the fact that $\ex{u_{i,\tau}\eta_{\tau,j}\mathbbm{1}_{|u_{i,\tau}\eta_{\tau,j}| \le \gamma} \mid \mathcal{F}_{\tau-1}} = -\ex{u_{i,\tau}\eta_{\tau,j}\mathbbm{1}_{|u_{i,\tau}\eta_{\tau,j}| > \gamma} \mid \mathcal{F}_{\tau-1}}$, we have
	\begin{lm}
		\begin{align*}
			\mathcal{T}_2 \le \sum_{i=1}^{m_n}\left( \frac{\sum_{\tau=1}^n |u_{i,\tau}|^{1+\alpha} c }{\gamma^{1+\alpha}}  + \left(\frac{\sum_{\tau=1}^n |u_{i,\tau}|^{1+\alpha} c }{\gamma^{1+\alpha}} \right)^2\right) \le \frac{m_n c n^{\frac{1-\alpha}{2} }}{\gamma^{1+\alpha}} + m_n\left(\frac{c n^{\frac{1-\alpha}{2} }}{\gamma^{1+\alpha}}\right)^2.
		\end{align*}
	\end{lm}
	The final result is obtained by substituting the value of $\gamma$, i.e., $(9m_n c)^{\frac{1}{1+\alpha}}n^{\frac{1-\alpha}{2(1+\alpha)} }$.
\end{proof}

\begin{lemma}
\label{lem:LSE}
	If $\gamma = (9m_n c)^{\frac{1}{1+\alpha}}n^{\frac{1-\alpha}{2(1+\alpha)} }$, $k = \lceil 24\ln\left(\frac{eT}{\delta}\right)\rceil$ and $\delta \in (0,1]$, then with probability at least $1-\delta$, uniformly over $n \in [N]$ 
	\begin{lm}
		\begin{align*}
			\norms{\tilde{V}_n^{-1}\tilde{\Phi}_n^Tf_n - \tilde{\theta}_{n,k^*} }_{\tilde{V}_n} \le 3\gamma.
		\end{align*}
	\end{lm}
	where $k^*$ is chosen based on the line $12$ of MoMA-GP-UCB algorithm. 
\end{lemma}
\begin{proof}
	This result can be proved by adapting the proof of Lemma 3 in~\cite{shao2018almost}. We define $b_j:= \mathbbm{1}_{\norm{\tilde{V}_n^{-1}\tilde{\Phi}_n^Tf_n - \tilde{\theta}_{n,j} }_{\tilde{V}_n} >  \gamma}$, $p_j := \mathbb{P}(b_j = 1)$ and $B_{\tilde{V}_n} (\tilde{V}_n^{-1}\tilde{\Phi}_n^Tf_n,\gamma ) := \{\theta: \norm{\theta - \tilde{V}_n^{-1}\tilde{\Phi}_n^Tf_n}_{\tilde{V}_n} \le \gamma \}$. 

	First, by Lemma~\ref{lem:LSE_j}, we have $p_j < \frac{1}{4}$. Then, by Hoeffding's inequality, we have 
	\begin{lm}
		\begin{align*}
			\mathbb{P}\left(\sum_{j=1}^k b_j \ge \frac{k}{3}\right) < \mathbb{P}\left( \sum_{j=1}^k b_j - p_j \ge \frac{k}{12}\right) \le e^{-\frac{k}{24} }.
		\end{align*}
	\end{lm}
	This means that with probability at least $1-e^{-\frac{k}{24} }$ more than $\frac{2}{3}$ of $\{\tilde{\theta}_{n,1},\ldots, \tilde{\theta}_{n,k} \}$ are within $B_{\tilde{V}_n} (\tilde{V}_n^{-1}\tilde{\Phi}_n^Tf_n,\gamma )$. When this is true, by the nature of $k^*$, we can directly have with probability at least $1-e^{-\frac{k}{24} }$, $\norms{\tilde{V}_n^{-1}\tilde{\Phi}_n^Tf_n - \tilde{\theta}_{n,k^*} }_{\tilde{V}_n} \le 3\gamma$. Therefore, by letting $k = \lceil 24\ln\left(\frac{eT}{\delta}\right)\rceil$, we have the desired result by union bound and $N = \lfloor T/k \rfloor$.
\end{proof}

Now, we are well-prepared to present the regret analysis of MoMA-GP-UCB.
\begin{proof}
	As pointed out before, the key is again a high probability confidence bound, which is now on $|f(x) - \tilde{\mu}_n(x)|$, uniformly over $n \in [N]$ and $x \in \mathcal{D}$. By Lemma~\ref{lem:CI_MoMA_pre}, we have when $E_{1,n}$ is true (i.e., $\varepsilon$-accurate), for any $x \in \mathcal{D}$, 
	\begin{lm}
		\begin{align*}
		|f(x) - \tilde{\mu}_n(x)| &\le B\left(1+\frac{1}{\sqrt{1-\varepsilon}}\right)\tilde{\sigma}_{n}(x) + \lambda^{-1/2}\norms{\tilde{V}_n^{-1}\tilde{\Phi}_n^Tf_n - \tilde{\theta}_{n,k^*} }_{\tilde{V}_n} \tilde{\sigma}_{n}(x).
		\end{align*}
	\end{lm}
	Then, combined with Lemma~\ref{lem:LSE}, we have that when the event $E_{1,n}$ is true, for any $\delta \in (0,1]$, when $k = \lceil 24\ln\left(\frac{eT}{\delta}\right)\rceil$, then with probability at least $1-\delta$, uniformly over $n \in [N]$ and $x \in \mathcal{D}$, we have 
	\begin{linenomath}
		\begin{align*}
		|f(x) - \tilde{\mu}_n(x)| &\le \left(B(1+\frac{1}{\sqrt{1-\varepsilon}})+\lambda^{-1/2}3\gamma \right) \tilde{\sigma}_{n}(x),
	\end{align*}
	\end{linenomath}
	where $\gamma = (9m_n c)^{\frac{1}{1+\alpha}}n^{\frac{1-\alpha}{2(1+\alpha)} }$. Since the event $E_{1,n}$ is true for all $n \in [N]$ with probability at least $1-\delta$ from Lemma~\ref{lem:events}, by using a union bound, we obtain that for any $\delta \in (0,1]$, if we choose $k = \lceil 24\ln\left(\frac{2eT}{\delta}\right)\rceil$, then with probability at least $1-\delta$, uniformly over all $n \in [N]$ and $x \in \mathcal{D}$, 
	\begin{linenomath}
		\begin{align}
		|f(x) - \tilde{\mu}_n(x)| &\le \beta_{n+1} \tilde{\sigma}_{n}(x),
	\end{align}
	\end{linenomath}
	where $\beta_{n+1}:=B(1+\frac{1}{\sqrt{1-\varepsilon}})+\lambda^{-1/2}3\gamma$. 

	Now, by the result above and the choice of MoMA-GP-UCB, we have for any $\delta \in (0,1]$, when $k = \lceil 24\ln\left(\frac{2eT}{\delta}\right)\rceil$
	\begin{lm}
		\begin{align*}
			r_n = f(x^*) - f(x_n) \le 2\beta_n \tilde{\sigma}_{n}(x)
		\end{align*}
	\end{lm}
	holds with probability at least $1-\delta$. Then, we first bound the term $\beta_n$. From Lemma~\ref{lem:events}, given a $(\mathcal{G}_n)_{n\ge1}$ such that the event $E_{2,n}$ is true for all $n \in [N]$, we have $m_n \le 6\rho\left(1+\frac{1}{\lambda}\right)q\gamma_t$. This implies that 
	\begin{lm}
		\begin{align*}
			\beta_n = O\left(B(1+\frac{1}{\sqrt{1-\varepsilon}}) + \left(\frac{\rho^2}{\varepsilon^2} \ln \frac{T}{\delta}\right)^{\frac{1}{1+\alpha}}  \gamma_n^{\frac{1}{1+\alpha} }c^{\frac{1}{1+\alpha}} n^{\frac{1-\alpha}{2(1+\alpha)} } \right).
		\end{align*}
	\end{lm}
	Next, we can bound the $\sum_{n=1}^N \tilde{\sigma}_{n}(x)$ as follows given that the event $E_{1,t}$ is true for $n \in [N]$.
	\begin{lm}
		\begin{align*}
			\sum_{n=1}^N \tilde{\sigma}_{n}(x) \lep{a} \rho\sum_{n=1}^N {\sigma}_{n}(x) \ep{b} O(\rho \sqrt{N\gamma_N}),
		\end{align*}
	\end{lm}
	where (a) holds by the second part of Lemma~\ref{lem:CI_MoMA_pre};  (b) holds by Lemma~\ref{lem:sum_sigma}. Recall that by Lemma~\ref{lem:events}, with probability at least $1-\delta$, both $E_{1,n}$ and $E_{2,n}$ are true for all $n \in [N]$. With the use of the union bound, we have for any $\delta\in (0,1]$, when $k = \lceil 24\ln\left(\frac{4eT}{\delta}\right)\rceil$, with probability at least $1-\delta$, the cumulative regret is given by
	\begin{lm}
		\begin{align*}
			R_T  = k\sum_{n=1}^N r_n \le k\sum_{n=1}^N 2\beta_n \tilde{\sigma}_{n}(x)= O\left(\hat{B}\sqrt{\gamma_T T \ln \frac{T}{\delta}} + Z \ln \frac{T}{\delta}c^{\frac{1}{1+\alpha}}T^{\frac{1}{1+\alpha}}\gamma_T^{\frac{3+\alpha}{2(1+\alpha)}}\right),
		\end{align*}
	\end{lm}
	where $\hat{B}=\rho B(1+\frac{1}{\sqrt{1-\varepsilon}})$ and $Z = \left(\frac{\rho^{3+\alpha}}{\varepsilon^2}\right)^{\frac{1}{1+\alpha}}$.
\end{proof}

\section{Additional Experiment Results}
\begin{figure*}[h]\centering
		\begin{subfigure}[b]{0.32\textwidth}
			\includegraphics[scale=0.14]{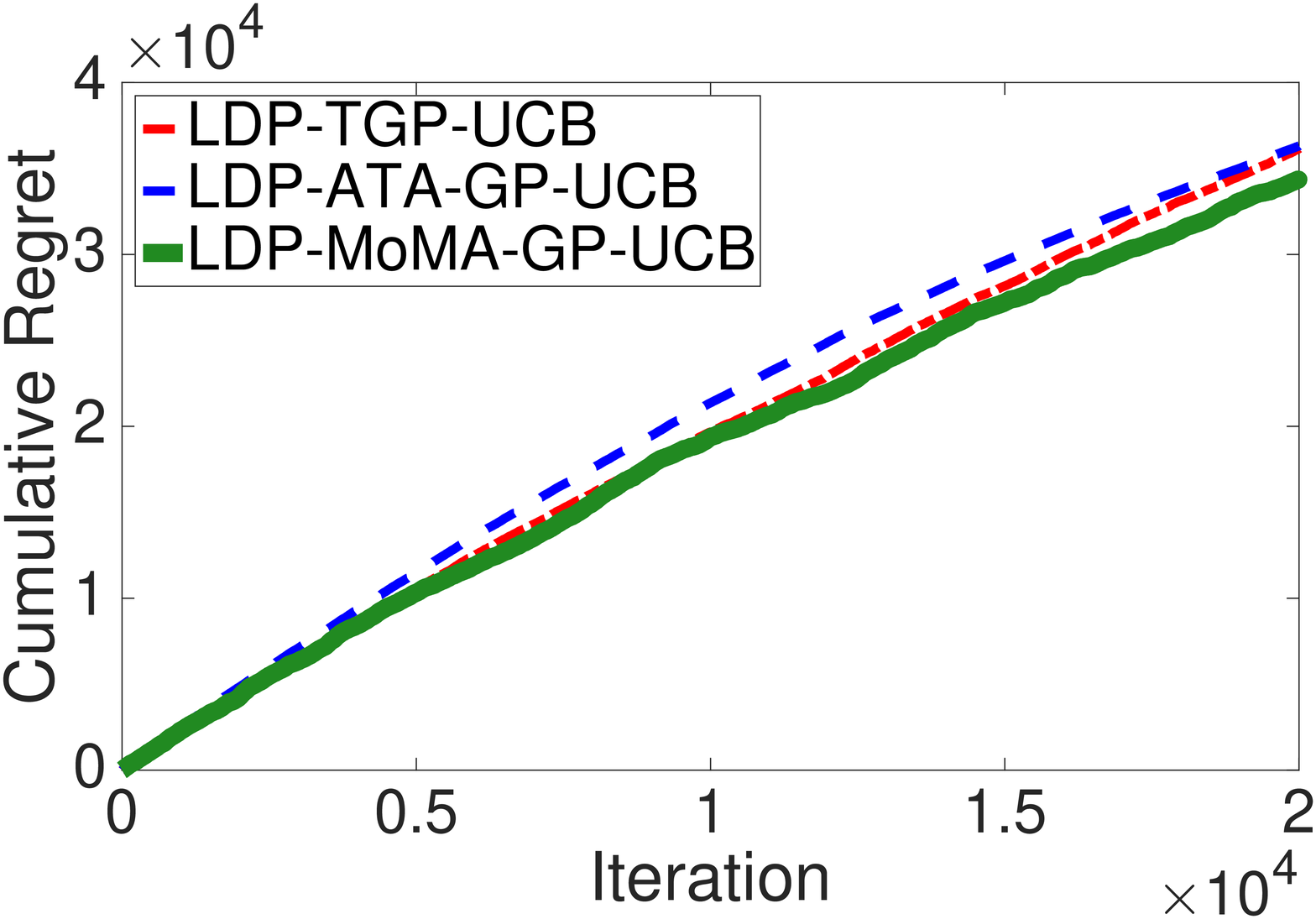}
			\caption{LDP, $\epsilon = 1$, Synthetic data, $k_{\text{SE} }$}
		\end{subfigure}\ \ 
		\begin{subfigure}[b]{0.32\textwidth}
			\includegraphics[scale=0.14]{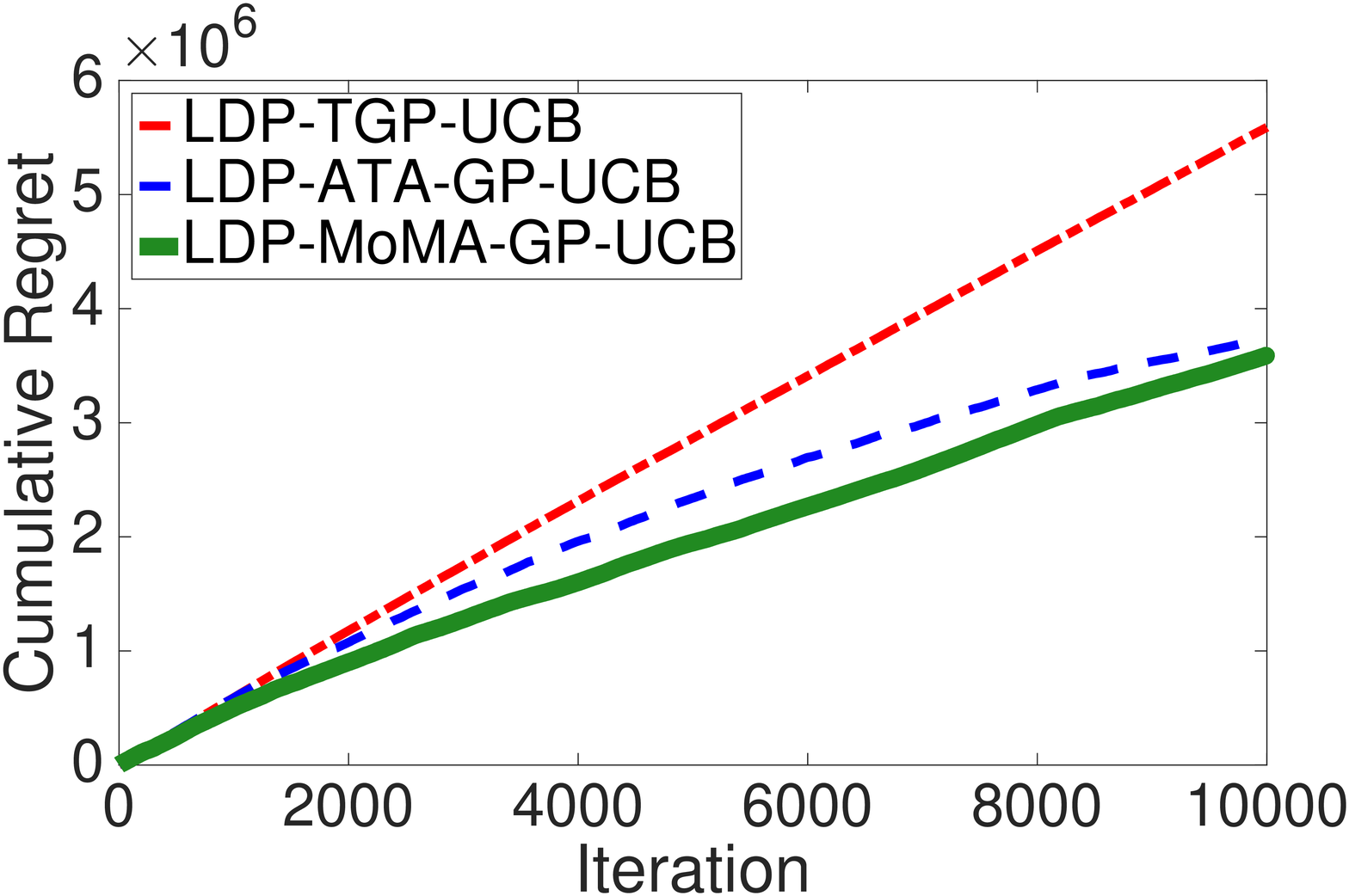}
			\caption{LDP, $\epsilon = 1$, Light sensor data}
		\end{subfigure}\ \ 
		\begin{subfigure}[b]{0.32\textwidth}
			\includegraphics[scale=0.14]{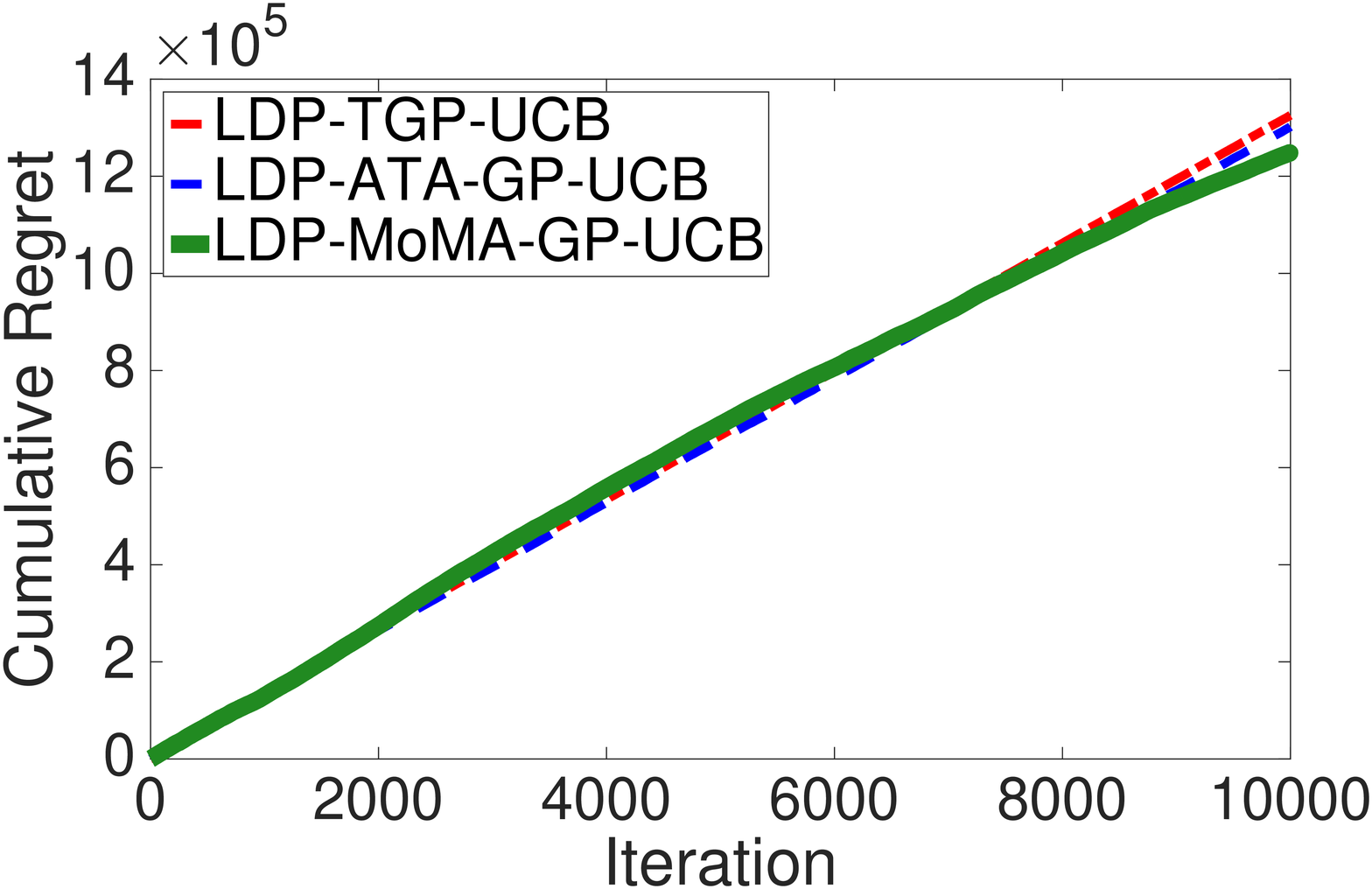}
			\caption{LDP, $\epsilon = 0.5$, Stock market data.}
		\end{subfigure}
		\caption{Cumulative regrets for three LDP algorithms under three new scenarios}\label{fig:regret_addition}
\end{figure*}

In this section, we conduct additional experiments to compare the performance of three LDP algorithms under different scenarios. For the synthetic case, we use $k_{\text{SE} }$ in this section. For the two real-world datasets, we use different values of $\epsilon$. From Fig.~\ref{fig:regret_addition}, we can see that LDP-MoMA-GP-UCB has better or competitive performance across different scenarios.
